\documentclass{article}

\PassOptionsToPackage{numbers, compress}{natbib}

\usepackage[preprint]{neurips_2024}

\usepackage[utf8]{inputenc} 
\usepackage[T1]{fontenc}    
\usepackage{hyperref}       
\usepackage{url}            
\usepackage{booktabs}       
\usepackage{amsfonts}       
\usepackage{nicefrac}       
\usepackage{microtype}      
\usepackage{xcolor}         
\usepackage{amsmath}
\usepackage{bbding}
\usepackage{makecell}
\usepackage{graphicx}
\usepackage{xspace}
\usepackage{multirow}
\usepackage{threeparttable}
\usepackage{xcolor}
\usepackage{colortbl}
\usepackage{adjustbox}
\usepackage{subfigure}
\usepackage{subcaption}
\usepackage{diagbox}
\usepackage{longtable}
\usepackage{rotating}

\newcommand{\ie}{\textit{i}.\textit{e}.}
\newcommand{\eg}{\textit{e}.\textit{g}.}

\title{Learning Robust 3D Representation from CLIP \\ via Dual Denoising}

\author{%
  Shuqing Luo\textsuperscript{1} \quad Bowen Qu\textsuperscript{1} \quad Wei Gao \textsuperscript{1} \\
\textsuperscript{1} School of Electronic and Computer Engineering, Peking University \\
\texttt{\{luoshuqing, bowenqu\}@stu.pku.edu.cn, gaowei262@pku.edu.cn} \\
}

\begin{document}

\maketitle

\begin{abstract}
In this paper, we explore a critical yet under-investigated issue: how to learn robust and well-generalized 3D representation from pre-trained vision language models such as CLIP. Previous works have demonstrated that cross-modal distillation can provide rich and useful knowledge for 3D data. However, like most deep learning models, the resultant 3D learning network is still vulnerable to adversarial attacks especially the iterative attack. In this work, we propose Dual Denoising, a novel framework for learning robust and well-generalized 3D representations from CLIP. It combines a denoising-based proxy task with a novel feature denoising network for 3D pre-training. Additionally, we propose utilizing parallel noise inference to enhance the generalization of point cloud features under cross domain settings. Experiments show that our model can effectively improve the representation learning performance and adversarial robustness of the 3D learning network under zero-shot settings without adversarial training. Our code is available at \url{https://github.com/luoshuqing2001/Dual_Denoising}.
\end{abstract}

\section{Introduction}

Recent years have witnessed the rapid development of vision-language pre-training, exemplified by the pioneering work of CLIP (\textbf{C}ross \textbf{L}anguage \textbf{I}mage \textbf{P}retraining)~\cite{radford2021learning} and the following works like FLIP~\cite{Li_2023_CVPR_FLIP}, SLIP~\cite{mu2022slip}, EVA-CLIP~\cite{sun2023eva}, which have significantly influenced downstream tasks and related research, propelling further investigations into 3D deep learning. In the pre-trained Vision-Language Models (VLMs) like CLIP, the input images and texts are encoded respectively to be projected to a shared latent space, which can be leveraged for tasks such as classification in an open-vocabulary manner. This suggests that projecting 3D point clouds into the same latent space could similarly enhance 3D learning. 

Recent efforts have attempted to align point clouds with images and texts using VLMs in various data spaces—from rendering 3D objects into RGB images~\cite{zhang2022pointclip, zhu2023pointclip} to employing depth images~\cite{Huang_2023_ICCV_CLIP2Point} or aligning high-level features~\cite{Zeng_2023_CVPR_CLIP2, Xue_2023_CVPR_ULIP, qi2023contrast, zhou2023uni3d}. This work focuses on developing a 3D encoder that not only projects point clouds into the shared pre-trained feature space but also improves generalization and adversarial robustness. Given that images alone cannot fully capture the geometry of 3D objects and 2D-based methods often fall short in performance, our research does not engage with these approaches.


Pre-trained VLMs can be distilled to learn 3D representations, which is usually associated with 3D pre-training~\cite{zhou2023uni3d, Xue_2023_CVPR_ULIP, xue2023ulip_2}. Through scaling up, it can achieve higher performance in zero-shot recognition tasks. Previous works~\cite{qi2023contrast} have demonstrated that the combination of self-supervised learning like mask reconstruction with knowledge distillation (also can be viewed as cross-modal contrastive learning) can improve the cross domain performance. In this work, we attempt to view this task from the perspective of learning adversarially robust representations. As a long-standing issue for deep learning, adversarial attacks can reduce the performance of deep neural networks by adding optimized tiny perturbations on the input. These perturbations are typically imperceptible to humans but can mislead the model to incorrect outputs. Previous works have proposed many talented solutions to alleviate this teaser, such as adversarial training~\cite{shafahi2019adversarial}, feature denoising~\cite{xie2019feature} and adversarial purification~\cite{yoon2021adversarial, nie2022diffusion}. However, simultaneously enhancing representation learning performance and adversarial robustness has proven challenging. In this work, we propose to train a \textit{Denoising Autoencoder} (DAE)~\cite{vincent2008extracting} on point cloud as a proxy task to learn robust and multi-modal features from CLIP. Experiments show that not only can our model surpass other methods with similar scale on zero-shot classification benchmark, but it is also more robust to 3D adversarial attacks, \ie, the robust accuracy of our method is higher than others under similar settings. 

\begin{figure}[!htbp]
\centering

\begin{minipage}[t]{0.99\linewidth}
\subfigure[Raw Point Cloud.]{
\centering
\includegraphics[width=0.31\linewidth, height=0.16\linewidth]{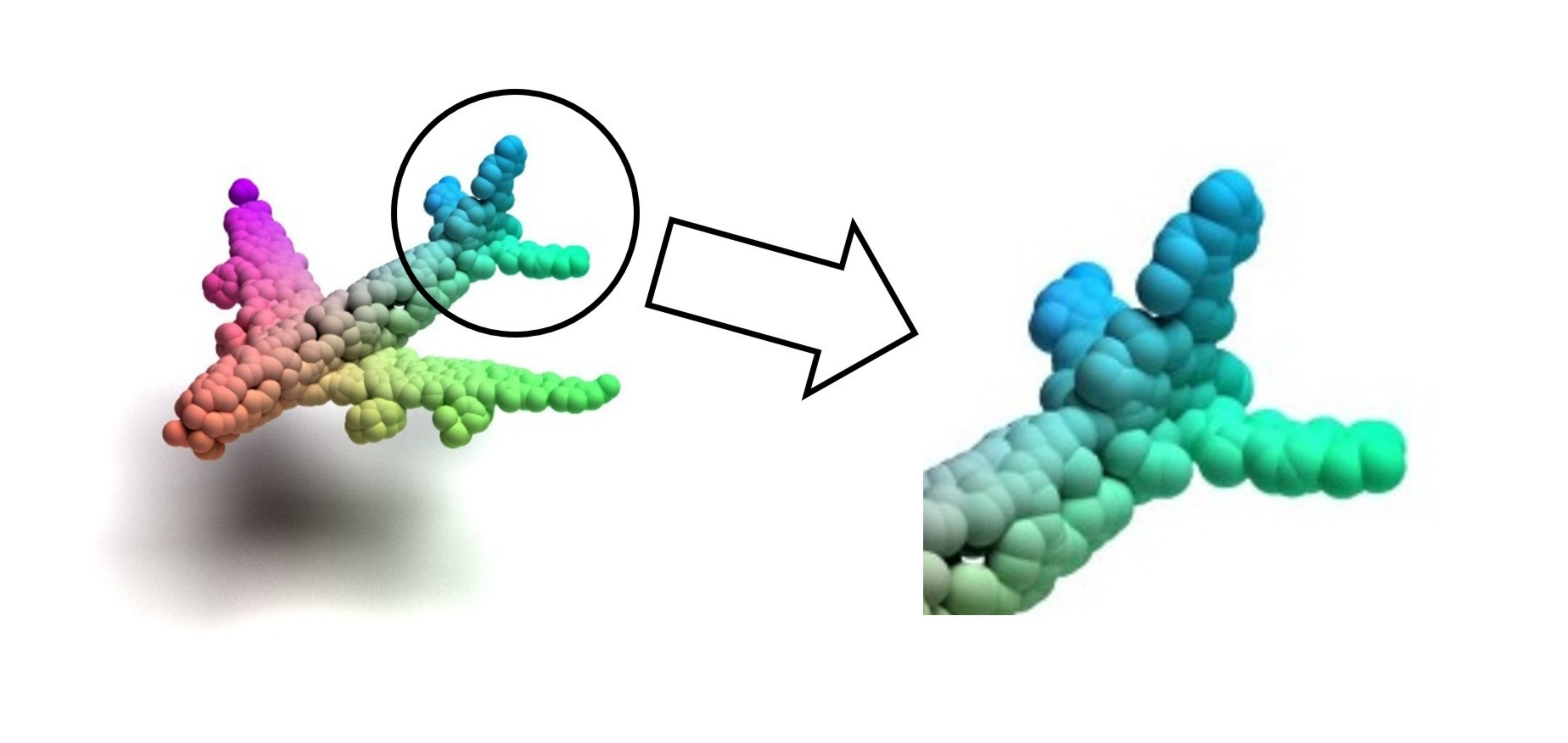}
}
\subfigure[Masked Point Cloud.]{
\centering
\includegraphics[width=0.31\linewidth, height=0.16\linewidth]{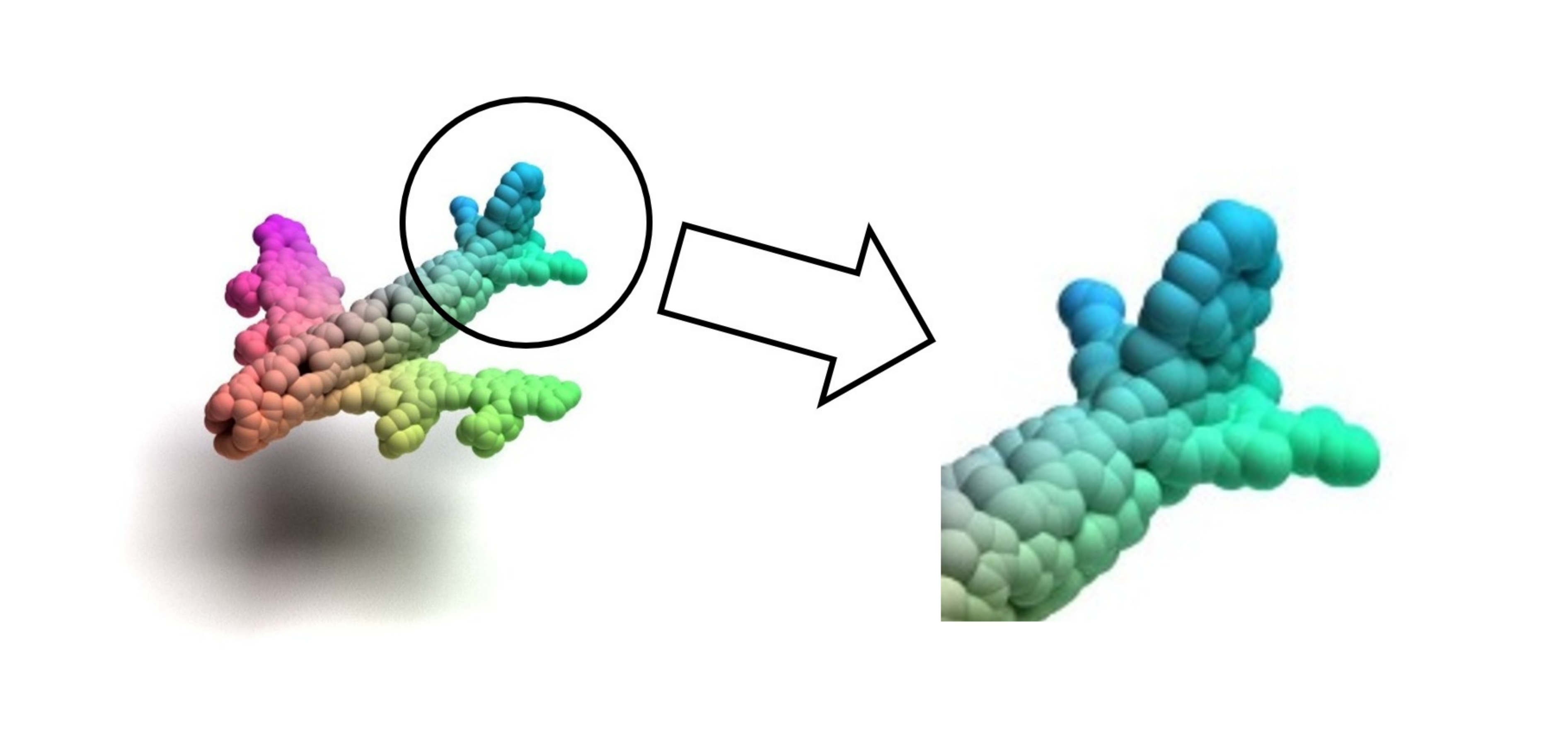}
}
\subfigure[Noised Point Cloud.]{
\centering
\includegraphics[width=0.31\linewidth, height=0.16\linewidth]{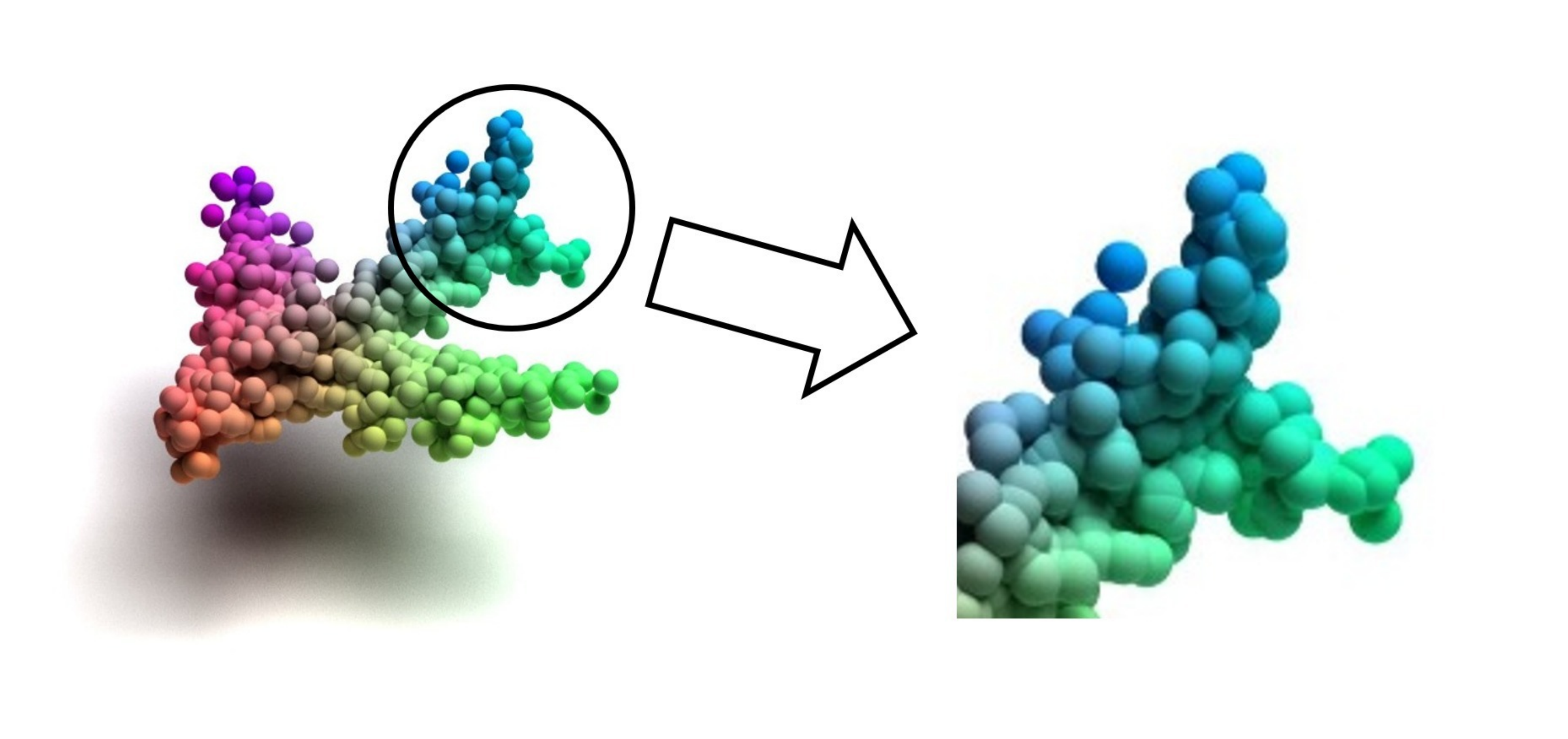}
}
\end{minipage}

\caption{\textbf{Visualization of raw point cloud, masked point cloud, and noised point cloud.} The raw point cloud contains 1024 points. For masked point cloud, we first compute 128 centroids using farthest point sampling (FPS) algorithm, then select 16 neighbors for each centroid using $K$ nearest neighbor (kNN) algorithm. We mask 80\% of the clusters and visualize the remained points. For noised point cloud, we diffuse the input by $x_t=x_0+\sigma_t\epsilon$, where $\{\sigma_t\}_{t=0}^{T-1}$ is a linear schedule from 0 to $s=0.08$. We choose $t=600$ and $T=1000$.}
\label{Fig:Clean_Masked_Noised}
\end{figure}

Masked autoencoders are often used in 3D pre-training. However, we present a comparison in Figure~\ref{Fig:Clean_Masked_Noised} to show that mask reconstruction may not be the optimal choice for point cloud pre-training. We take the point cloud representation of an airplane as an example. We can see that the tail of it in the raw point cloud and masked point cloud does not have much difference even with a masking ratio of 80\%. Therefore, learning to reconstruct the masked point tokens does not have too much difference from reconstructing all of the original point tokens themselves, \ie, setting mask ratio to 0. We also confirm it in the experiment results. However, in Figure~\ref{Fig:Clean_Masked_Noised}(c), we can find that the tail of the noised plane has been obviously disturbed, which makes learning to reconstruct the clean point cloud non-trivial. We also present in the experiments that the proxy task of denoising can largely improve the model generalization in zero-shot recognition tasks. 

To learn robust and well-generalized representations, we designed a novel dual denoising architecture for our model, which is inspired by diffusion-based generative models like DDPM~\cite{ho2020denoising} and the relevant research on its representation learning ability~\cite{chen2024deconstructing}. Specifically, our model is composed of two branches, one performs denoising on the point cloud as a proxy task while the other transforms another input standard gaussian noise to the corresponding CLIP features. These two branches are densely coupled via cross-attention modules. The noise added on point cloud follows a pre-defined scheduler similar to DDPM, while the noise used for feature branch is a gaussian noise with fixed variance. We also propose a test time augmentation (TTA) strategy to further improve the generalization performance by parallel noise inference.

The contributions of our model can be outlined as follows:
\begin{enumerate}
\item[(1)] We propose to learn robust 3D representation via distillating CLIP from the perspective of adversarial machine learning. Our method can surpass previous methods with similar scale and configuration with regard to model and training dataset on zero-shot point cloud classification benchmark.
\item[(2)] We propose \textit{Dual Denoising} as a robust cross-modal distillation framework for 3D point cloud, which shows better adversarial robustness under zero-shot settings. We also extend the 3D adversarial attack algorithms to multi-modal settings. Experiments demonstrate the effectiveness of these two novel designs.
\item[(3)] We propose parallel noise inference to further improve the cross domain representation learning performance of our model, which does not require too much computation budget. Experiments also demonstrate the effectiveness of this strategy.
\end{enumerate}

\vspace{-0.4cm}
\section{Related Work}
\paragraph{Contrastive Language Image Pre-Training.} The pioneering work CLIP~\cite{radford2021learning} from OpenAI provides a new paradigm for multi-modal learning by projecting vision and language inputs to a shared latent space respectively. Classification is implemented in a retrieval-based manner. Given input image $\boldsymbol{x}$ and label set $\boldsymbol{y}=\{y_i|i=1,\dots,N\}$, we first compute the $l_2$-normalized image feature $\boldsymbol{f}_v(\boldsymbol{x})$ and text feature $\boldsymbol{f}_l(\boldsymbol{y})$, then get the logits by
\begin{equation}
\label{eq:CLIP}
    \boldsymbol{logits}=e^t\cdot\langle \boldsymbol{f}_v(\boldsymbol{x}), \boldsymbol{f}_l(\boldsymbol{y})\rangle,
\end{equation}
where $t$ is a pre-defined constant. Then we can get the probability of each class by a \textit{Softmax} operation on the logits. Notice that the labels are given as texts rather than integers, thus it can empower the model to inference in an open-vocabulary manner. Prompt engineering is also necessary for CLIP to get better cross domain performance. Some following works try to enhance the cross domain generalization by improving the training strategy. FLIP~\cite{Li_2023_CVPR_FLIP} propose to accelerate and scale the training of CLIP via masking. SLIP~\cite{mu2022slip} propose to combine self-supervised learning and CLIP pre-training. EVA-CLIP~\cite{sun2023eva} incorporates new techniques for representation learning, optimization, and augmentation to enable it achieving better performance than CLIP with equal number of parameters and smaller training costs. SigLIP~\cite{zhai2023sigmoid} propose to replace \textit{Softmax} in CLIP with \textit{Sigmoid} to improve the efficiency of large-scale distributed pre-training, as computing \textit{Sigmoid} loss does not need to traverse the whole batch. 

\paragraph{Adversarial Robustness of 3D Deep Learning Models.} Deep learning has been widely adapted on point cloud data~\cite{qi2017pointnet, qi2017pointnet++, guo2021pct, zhao2021point}, and previous works have also demonstrated that the vulnerability of deep learning model under adversarial attacks~\cite{goodfellow2014explaining, madry2018towards} still exists in 3D models~\cite{xiang2019generating, hamdi2020advpc}. To tackle this problem, researchers have proposed to use more robust network~\cite{zhou2019dup}, integrate adversarial training~\cite{sun2020adversarial} and purification on the input~\cite{sun2023critical}. In this paper, we focus on designing pre-training algorithm on point cloud that is more robust on downstream tasks, without the aid of adversarial training.

\paragraph{Denosing Methods for Representation Learning.} Denoising Diffusion Probablistic Models (DDPM)~\cite{ho2020denoising} have made great contributions for generative tasks~\cite{dhariwal2021diffusion, rombach2022high, peebles2023scalable}, which also inspire the research on representation learning~\cite{xiang2023denoising, chen2024deconstructing}. The pioneering work \textit{l-DAE}~\cite{chen2024deconstructing} provides a through study on the components of DDPM to explore their contributes on representation learning, respectively. It finds that the latent space provided by the \textit{off-the-shelf} tokenizer, the denoising process and the prediction of clean input are beneficial for representation learning, while the label condition, the noise schedule and the input scaling are unnecessary.

\section{Our Method}

\begin{figure}[!htbp]
\centering
\includegraphics[scale=0.09]{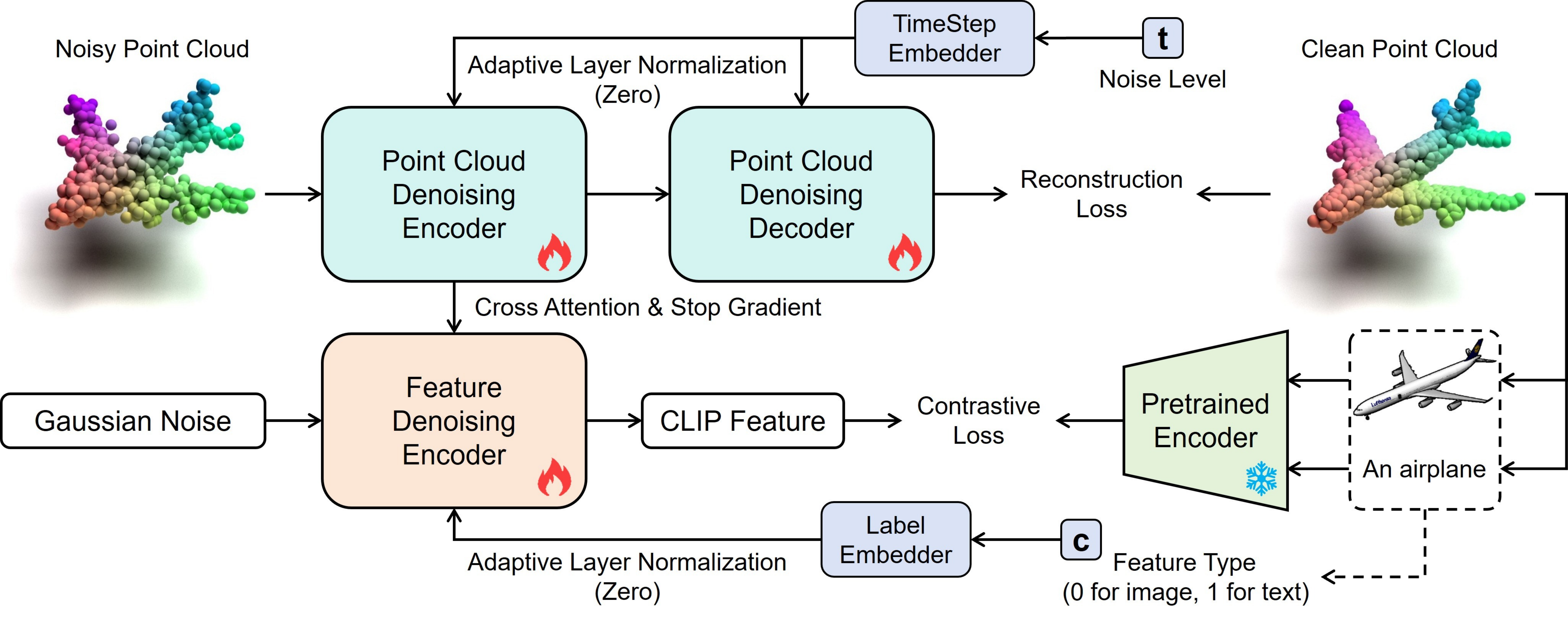}
\caption{\textbf{Pipeline of our \textit{Dual Denoising} algorithm.} The upper branch is PointDAE, performing as a proxy task during pre-training. The lower branch is feature denoising, gradually transforming a gaussian noise to CLIP feature under the guidance from the upper branch.}
\label{fig:Pipeline}
\end{figure}

The overview of our robust distillation algorithm is shown in Figure \ref{fig:Pipeline}. It is composed of two branches. The upper branch, named as Point Denoising AutoEncoder (PointDAE), conducts denoising for point cloud following a pre-defined noise scheduler, performs as a proxy task. The lower branch is named as feature denosing. It converts the input standard gaussian noise to CLIP feature hierarchically under the guidance of point cloud features from the upper branch. The two branches are densely connected by multiple cross-attention modules. A stop gradient operation~\cite{chen2021exploring} is implemented on them during pre-training to avoid representation collapse. Moreover, we propose a new test time augmentation strategy, which can improve the performance of cross domain generalization by parallel noise inference.

\subsection{PointDAE}

PointDAE performs as a proxy task for pre-training. It leverages a ViT~\cite{dosovitskiy2020image}-like encoder and decoder architecture to predict clean point cloud from the noisy input. Similar to classical Denoising Diffusion Models (DDM), we compute noisy point cloud from the clean data $z_0$ via a diffusion process. At time step $t$, the noised data $z_t$ is derived from:
\begin{equation}
\label{eq:Diffusion}
    z_t=\gamma_tz_0+\sigma_t\epsilon,
\end{equation}
where $\epsilon\sim\mathcal{N}(0,\mathbf{I})$ is a standard gaussian noise. As input scaling and noise schedule is unessential for representation learning~\cite{chen2024deconstructing}, we set $\gamma_t\equiv 1$ and $\sigma_t$ as a simple linear schedule from $0$ to $s$. The noise time step $t$ is further embedded as a guidance for the point cloud learning backbone, using Adaptive Layer Normalization Zero (AdaLN-Zero)~\cite{peebles2023scalable}. 

\begin{figure}[!htbp]
\centering
\includegraphics[scale=0.09]{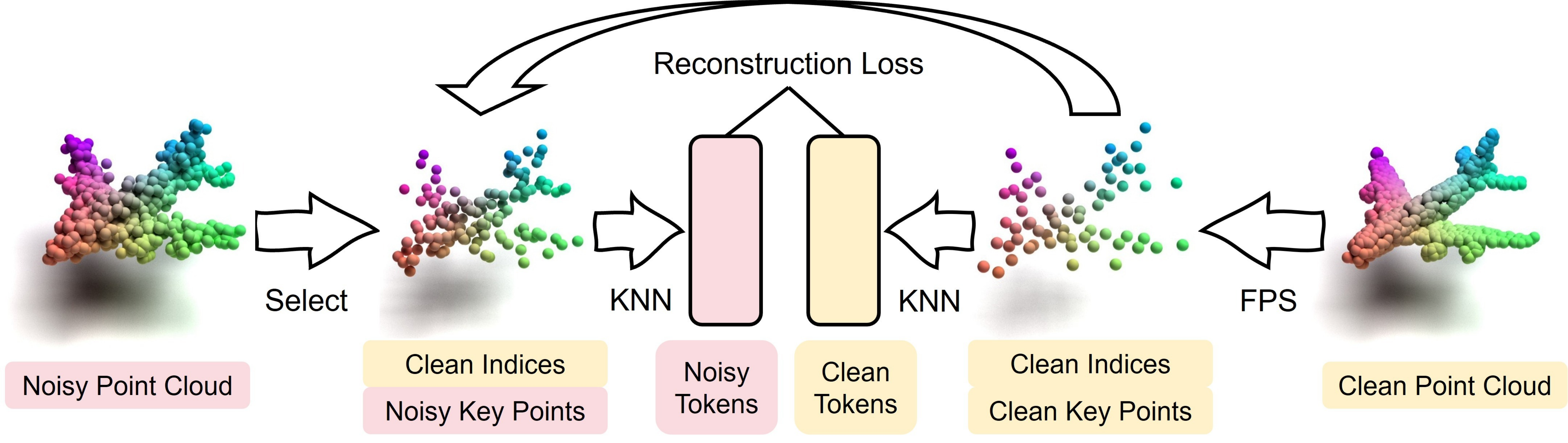}
\caption{\textbf{Implementation of PointDAE during training.} Point cloud is tokenized to fit the ViT architecture. We first using farthest point sampling (FPS) on the clean data to get a representative subset of indices. Then we compute the clean key point subset and noisy one under the same indices. Next we conduct $k$ nearest neighbor (kNN) to get the noisy point tokens and clean tokens respectively. Reconstruction loss like chamfer distance is used between them.}
\label{fig:PointDAE}
\end{figure}

Although the latent space provided by the \textit{off-the-shelf} tokenizer is important for DDM inspired representation learning~\cite{chen2024deconstructing}, it is difficult to project point cloud to some latent space and add noise on it following some schedule. This is because point cloud tokens are overlapped between each other, and the basic element in each token (point) lies in a disordered space. To adapt the mechanism of DDM to point cloud representation learning, we designed a simple yet efficient training scheme, shown in Figure \ref{fig:PointDAE}. We first add noise on the $xyz$-coordinates of the input point cloud to perturb the whole point set. To transform it into tokens, we conduct farthest point sampling (FPS) on the clean data to get a subset composed of key point indices. Then we select the noisy key point subset from the noised data with these indices. Noisy tokens and clean tokens can be obtained respectively afterwards, using $k$ nearest neighbor (kNN) algorithm. During pre-training, we compute reconstruction loss for each token and average them, denoted as $\mathcal{L}_{p}^t$. To make the model learn more from the cleaner data so that to better align with other modalities, we empirically set the loss weight as $\lambda_t=1/(1+\sigma_t^2)$.

\begin{figure}[!htbp]
\centering
\includegraphics[scale=0.084]{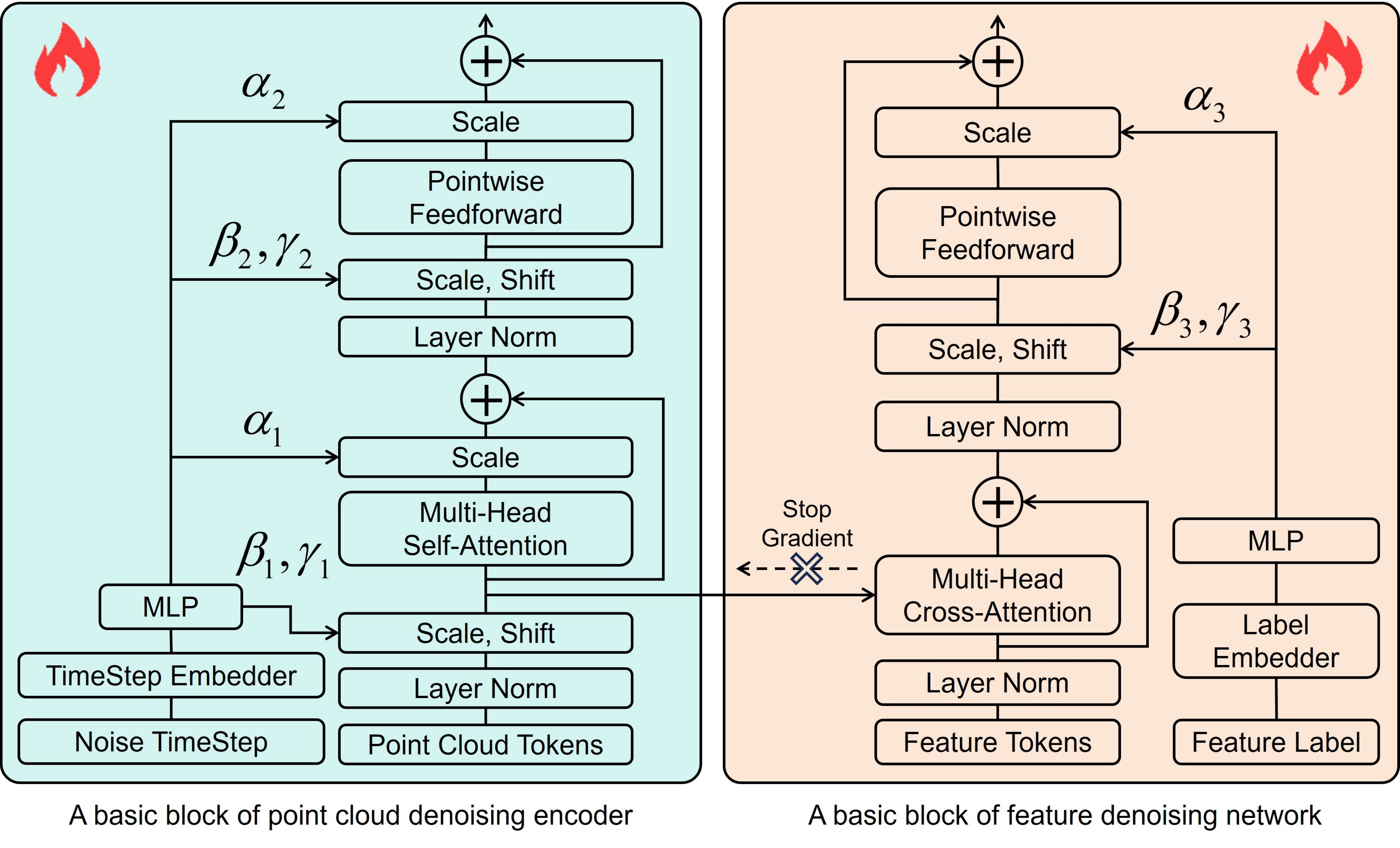}
\caption{\textbf{The details of a basic block in PointDAE (left) and feature denoising network (right).} Notice that the structure of point cloud denoising decoder is similar to the encoder, only without the cross-attention connection with feature denoising branch. Point cloud tokens perform as $K$ and $V$ while feature tokens perform as $Q$ in the cross attention module. A stop gradient operation is jointly used during pre-training to avoid representation collapse.}
\label{fig:basic_block}
\end{figure}

\subsection{Feature Denoising}

The feature denoising network gradually transforms a standard gaussian noise to high-level features, guided by point cloud features from each layer in the point denoising encoder. We train this part to learn $k$ kinds of feature from CLIP, \eg, image feature or text feature. This part is pre-trained using contrastive learning to align with CLIP, and the corresponding loss function is denoted as $\mathcal{L}_{f}$. We use a weighted combination of point token reconstruction loss at time step $\mathcal{L}_{p}^t$ and feature contrastive loss $\mathcal{L}_{f}$ for our robust pre-training, with a loss weight $\alpha$ for $\mathcal{L}_f$:
\begin{equation}
    \mathcal{L}_{train}=\lambda_t\cdot\mathcal{L}_{p}^t+\alpha\cdot\mathcal{L}_{f}.
\end{equation}

\subsection{Basic Blocks}

The detailed structure of basic blocks in PointDAE and feature denoising network is shown in Figure~\ref{fig:basic_block} as the left and right one. In PointDAE, we need to modulate the noise time step into the basic modules. Following DiT~\cite{peebles2023scalable}, we use a timestep embedder on it, followed by a multi-layer perception (MLP) to regress $\alpha$, $\beta$ and $\gamma$. Since this AdaLN-Zero module provides a guidance for the backbone network that may not be beneficial for representation learning, we explore to use a time step merging strategy to weaken this guidance, making this proxy task more challenging. As a standard DDM usually use a total number of $1000$ noise time steps, we use a reduced total number as input for the neural network while keeping the noise schedule unchanged. Denoting the merging interval as $\Delta$, the weakened time step for $t$ is obtained by $t^{\prime}=\lfloor t/\Delta\rfloor$. When we set $\Delta$ as a relatively large number such as $100$ and use $t^{\prime}$ as the model input, we can train the model to be more robust to the noise level so that to learn better representations.

The category of feature is denoted as integers $0,\dots,k-1$, which is embedded with a learnable embedding layer and projected with a MLP to regress $\alpha, \beta$ and $\gamma$, performing as AdaLN-Zero~\cite{peebles2023scalable} on the feedforward network. The cross attention context comes from PointDAE. To avoid over-fitting and representation collapse, a stop gradient operation~\cite{chen2021exploring} is necessary in the pre-training stage. 

\subsection{Parallel Noise Inference}

We propose to use multiple noise as input for inference parallelly as a test time augmentation strategy. Denoting our model as $y_i^t=f_i(x, t, \epsilon_1^t, \epsilon_2)$, where $y$ is the predicted CLIP feature, $x$ is the input point cloud, $t\in\{0,\dots,T-1\}$ is the diffuse time step, $i\in\{0,\dots,k-1\}$ is the feature type and $\epsilon_1^t, \epsilon_2$ are the standard gaussian noise with the same shape as $x, y$. Since we take denoising as the proxy task, the input of the model is reasonable to be a noised point cloud. We propose to inference in a parallel manner, aggregating $N$ paths: 
\begin{equation}
\label{eq:TTA}
    y_i^t=\frac{1}{N}\cdot\sum\limits_{j=1}^{N}f_i(x, t, \epsilon_{j,1}^t, \epsilon_{j,2}),
\end{equation}
where $\{\epsilon_{j,1}^t\}_{j=1}^N$ and $\{\epsilon_{j,2}\}_{j=1}^N$ are independently sampled from $\mathcal{N}(0,\mathbf{I})$. After that, we can use $\{y_i^t\}$ for knowledge ensemble, \ie, taking weighted sum of these features to boost the performance in cross domain tasks.

\section{Experiments}

\subsection{Implementation}

So far, there has been quite a few solutions on learning 3D representation from pre-trained VLMs with various experiment settings. For example, the choice of pre-training dataset includes ShapeNet~\cite{chang2015shapenet} and Objaverse~\cite{deitke2023objaverse} with a size of 50K+ and 800K+. The prompt templates include hand-crafted~\cite{zhang2022pointclip, qi2023contrast} and synthesized, with the help of LLMs~\cite{zhu2023pointclip} or multi-modal LLMs~\cite{xue2023ulip_2, qi2024shapellm}. The scale of model architecture also varies from millions to billions. To make a fair comparison with existing methods, we take the basic experiment setting similar to ReCon~\cite{qi2023contrast}, including: \textbf{(1) Dataset:} we take ShapeNet, the most commonly used dataset for 3D pre-training. \textbf{(2) Prompt:} we follow PointCLIP~\cite{zhang2022pointclip} and use the hand-crafted templates. \textbf{(3) Model:} we use the standard plain transformer~\cite{vaswani2017attention} encoder blocks with dimension 384 and a tiny PointNet patch embedding module to learn 3D tokens. The PointDAE encoder contains 12 blocks and decoder contains 4 blocks. We use the Vision Transformer (ViT-B)~\cite{dosovitskiy2020image} and text encoder from CLIP~\cite{radford2021learning} as the vision teacher and language teacher, respectively. The image and text teacher encoders are frozen during pre-training, using Smooth $l_1$-based positive-only distillation loss~\cite{chen2021exploring}. PointDAE uses a reconstruction loss based on $l_2$ Chamfer-Distance. All the experiments are conducted on a single NVIDIA GeForce RTX 3090.

\subsection{Zero-Shot 3D Object Recognition}

Since we align the 3D feature with pre-trained CLIP feature space, our model has a competitive zero-shot capability. We use multiple dataset for zero-shot evaluation, following the previous benchmark~\cite{zhu2023pointclip}. The evaluation datasets include real-world object recognition dataset ScanObjectNN and synthetic object dataset ModelNet. ScanObjectNN~\cite{uy2019revisiting} is one of the most common and challenging 3D datasets containing $\sim 15$ K real-world objects from 15 categories. We take 3 splits of it, including OBJ\_ONLY, OBJ\_BG, and PB\_T50\_RS. ModelNet~\cite{wu20153d} is also a commonly used 3D dataset, containing $\sim 12$ K CAD objects of 40 (ModelNet40) or 10 (ModelNet10) categories. We implement evaluation on both datasets. Following the zero-shot principle, we directly test the classification performance on the full test set without learning from the training set. We compare existing methods under their best settings to fully achieve their performance, following PointCLIP v2~\cite{zhu2023pointclip}. For fair comparison, we do not compare with methods that scale up on model size or pre-training dataset like ULIP-2\cite{xue2023ulip_2} and Uni3D~\cite{zhou2023uni3d}, or methods that adopt external knowledge from LLMs like ShapeLLM~\cite{qi2024shapellm}. The zero-shot 3D object classification results are shown in Figure~\ref{table:zero_shot_classification}. We surpass almost all of the previous methods with similar configuration or scale. For ModelNet, we achieve $79.5\%$ accuracy on ModelNet10 and $69.0\%$ accuracy on ModelNet40, with an improvement of $3.9\%$ and $4.8\%$. For ScanObjectNN, we achieve $52.7$ accuracy on OBJ\_ONLY, $48.7$ accuracy on OBJ\_BG, and $39.8$ accuracy on PB\_T50\_RS, with an improvement of $2.6\%$, $4.5\%$, and $4.4\%$, respectively. The best performance of our method is obtained with $N=8, t=600$ on ModelNet and $N=8, t=100$ on ScanObjectNN. More details are shown in Figure~\ref{tab:Ablation_Parallel}. 

\begin{table*}[!htbp]
\centering
\begin{adjustbox}{width=1.0\linewidth}
	\begin{tabular}{lccccccc}
	\toprule
		Method & Pre-train Dataset & Teacher Model &ModelNet10 & ModelNet40 &S-OBJ\_ONLY &S-OBJ\_BG &S-PB\_T50\_RS \\
		\cmidrule(lr){1-1}
		\cmidrule(lr){2-3} 
		\cmidrule(lr){4-8} 
            CLIP2Point~\cite{Huang_2023_ICCV_CLIP2Point} & ShapeNet & CLIP & $66.6$ & $49.4$ & $35.5$ &$30.5$ & $23.3$ \\
            PointCLIP~\cite{zhang2022pointclip} \vspace{0.05cm} & - & - &$30.2$ & $23.8$  &$21.3$ &$19.3$ & $15.4$ \\
            PointCLIP V2~\cite{zhu2023pointclip} \vspace{0.05cm} & - & - &${73.1}$ &${64.2}$  &${50.1}$ &${41.2}$ &${35.4}$\\
            ULIP~\cite{Xue_2023_CVPR_ULIP} & ShapeNet & SLIP & $72.8$ &${60.4}$  &${49.9}$ & $44.2$ & $27.2$ \\
            ReCon~\cite{qi2023contrast} & ShapeNet & CLIP &${75.6}$ &${61.7}$  &${43.7}$ &${38.6}$ &${28.6}$\\
            \textbf{Ours} & ShapeNet & CLIP &$\textbf{79.5}$ &$\textbf{69.0}$  &$\textbf{52.7}$ &$\textbf{48.7}$ &$\textbf{39.8}$\\
     \textit{Improvement} &&&\textcolor{blue}{$+3.9$}&\textcolor{blue}{$+4.8$}&\textcolor{blue}{$+2.6$}&\textcolor{blue}{$+4.5$}&\textcolor{blue}{$+4.4$}\\
	\bottomrule
	\end{tabular}
\end{adjustbox}
\caption{\textbf{Zero-shot 3D classification accuracy (\%) on ModelNet10, ModelNet40 and ScanObjectNN.} We report the performance of other methods with their \textbf{\textit{best-performing settings, \eg, visual encoder, projected view number, and textual input}}.}
\label{table:zero_shot_classification}
\vspace{-0.1cm}
\end{table*}

\subsection{Adversarial Robustness under Zero-Shot Settings}

\begin{figure}[!htbp]
\centering

\begin{minipage}[t]{1.0\linewidth}
\subfigure[Adversarial Robustness on ModelNet40.]{
\centering
\includegraphics[width=0.484\linewidth, height=0.28\linewidth]{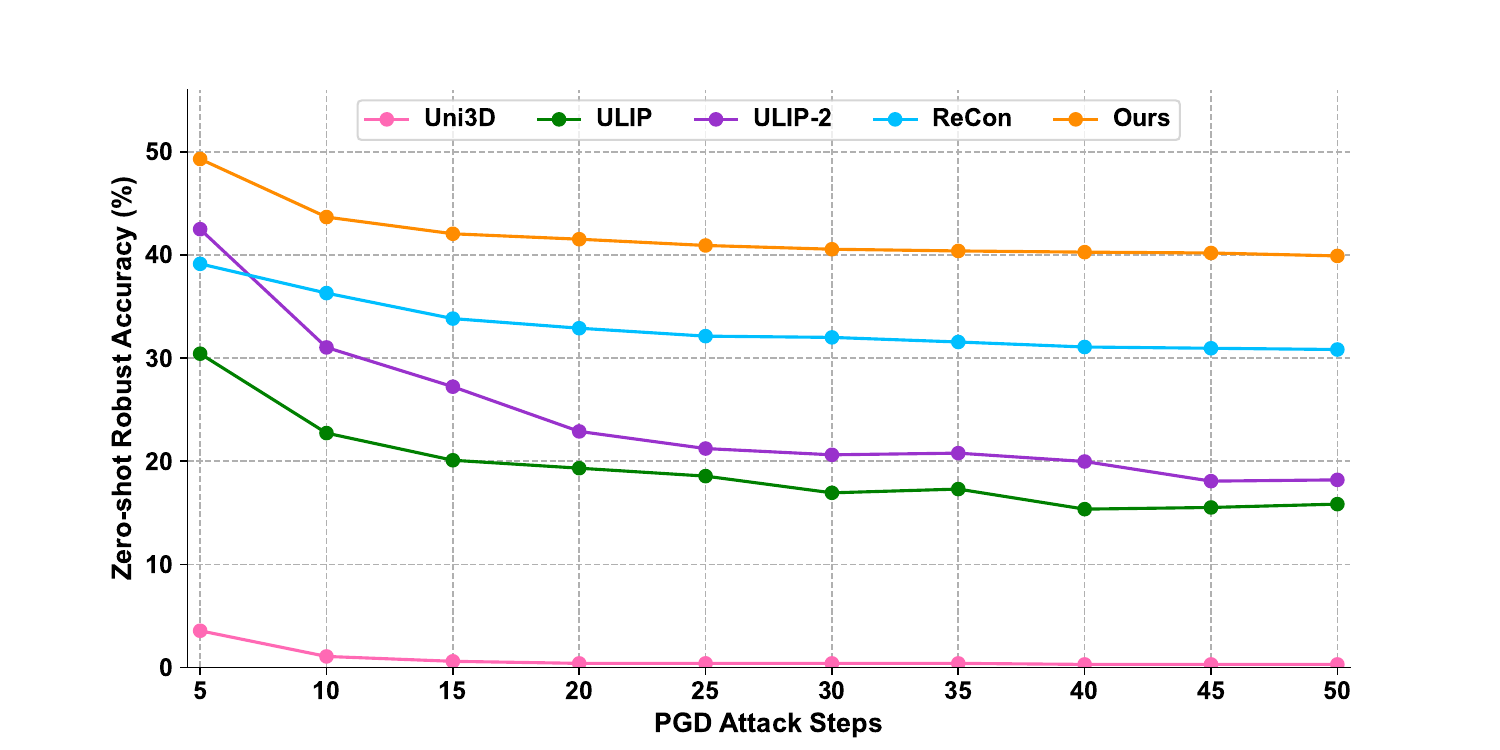}
}
\subfigure[Adversarial Robustness on ScanObjectNN.]{
\centering
\includegraphics[width=0.484\linewidth, height=0.28\linewidth]{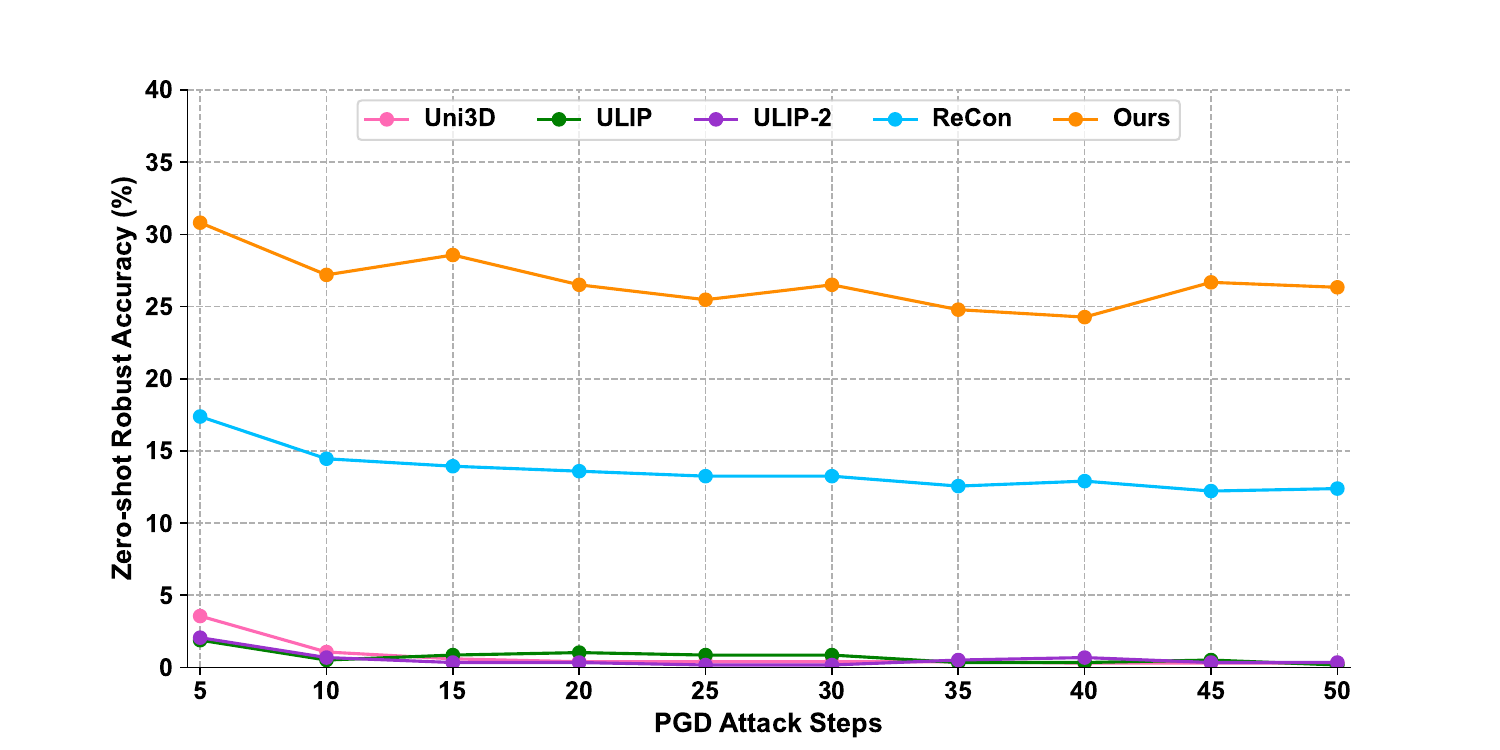}
}
\end{minipage}

\caption{\textbf{Visualization of adversarial robustness under PGD attack on ModelNet40 and ScanObjectNN.} We use ModelNet40 test set and OBJ\_ONLY test set, respectively. }
\label{Fig:Adv_PGD_Steps}
\end{figure}

We extend 3D adversarial attack algorithms in standard classification to zero-shot classification, and evaluate existing cross modal distillation algorithms for point cloud under the same setting. The gradient-based 3D adversarial attack algorithms can largely reduce the performance of the 3D learning model by adding slight perturbation on the input. Generally speaking, 3D adversarial example is computed by raising the \textit{logit} value of a target while minimizing the perturbation on the input through optimization~\cite{xiang2019generating}. In this way, we can change the model output while keeping the input point cloud almost unchanged. To extend previous methods to CLIP-like zero-shot classification task, we only need to change the \textit{logits} into cosine similarity in Eq~\ref{eq:CLIP}. We choose the first candidate (the element with 2nd highest similarity value) as the target, and conduct targeted adversarial attack~\cite{xiang2019generating}. We use iterative attack algorithms, including IFGM~\cite{ding2023cap}, PGD~\cite{sun2021adversarially} and C\&W Perturb~\cite{xiang2019generating}, as they have stronger attack capacity. Please see Appendix~\ref{appendix} for more details of these algorithms. Experiment results are shown in Table~\ref{tab:benchmark_adv}. We investigate current methods under these attacks, including ReCon~\cite{qi2023contrast}, ULIP~\cite{Xue_2023_CVPR_ULIP}, ULIP-2~\cite{xue2023ulip_2}, and Uni3D~\cite{zhou2023uni3d}. $\epsilon$ is used for gradient-based optimization, where larger one means larger degree of perturbation. We implement IFGM and PGD for 50 steps, while conduct C\&W Perturb attack for 10-step binary search and 500 iterations of optimization in each binary search to find the adversarial examples. From Table~\ref{tab:benchmark_adv} we can find that our method is more robust under adversarial attacks. We also visualize these methods under PGD attack under different optimization steps with $\epsilon=0.01$, shown in Figure~\ref{Fig:Adv_PGD_Steps}. This also shows the robustness of our method. Notice that we also compare with the methods that scaled up like Uni3D and ULIP-2. When taking adversarial examples as input, our model performs better than them.

\begin{table*}[!htbp]
\centering
\begin{adjustbox}{width=0.98\linewidth}
\begin{tabular}{lcccccccc}
\toprule
\multirow{2}{*}{Method}&\multicolumn{8}{c}{\makecell[c]{Adversarial robustness on zero-shot 3D classification task on ModelNet40 test dataset}} \\
\cmidrule(lr){2-9}
& \text{ReCon}~\cite{qi2023contrast} & \text{ULIP}~\cite{Xue_2023_CVPR_ULIP} & \text{ULIP-2}~\cite{xue2023ulip_2} & \text{Uni3D}~\cite{zhou2023uni3d} & \makecell{Ours \\ ($N$=8, $t$=100)} & \makecell{Ours \\ ($N$=8, $t$=500)} & \makecell{Ours \\ ($N$=8, $t$=900)} & \makecell{Ours \\ ($N$=16, $t$=500)}\\
\midrule
Clean point cloud & 61.7 & 60.3 & 75.6 & 86.3 & 68.4 & 68.8 & 68.3 & 68.6 \\
\midrule
IFGM ($\epsilon=0.01$) & 27.5 & 17.8 & 29.1 & 0.4 & 33.5 & 38.5 & 40.6 & \textbf{42.3} \\
IFGM ($\epsilon=0.025$) & 16.9 & 6.0 & 8.8 & 0.2 & 9.7 & \textbf{20.7} & 20.4 & 18.8 \\
IFGM ($\epsilon=0.05$) & 7.1 & 3.0 & 3.1 & 0.0 & 3.5 & 8.9 & \textbf{10.1} & 8.3 \\
IFGM ($\epsilon=0.075$) & 3.8 & 2.3 & 2.8 & 0.0 & 2.6 & \textbf{4.7} & 4.4 & 4.1 \\
\midrule
PGD ($\epsilon=0.01$) & 31.7 & 18.0 & 18.2 & 0.3 & 43.8 & $\mathbf{48.1}$ & 46.3 & 47.9 \\
PGD ($\epsilon=0.025$) & 13.3 & 5.0 & 5.3 & 0.0 & 24.9 & 31.8 & $\mathbf{31.9}$ & 31.8 \\
PGD ($\epsilon=0.05$) & 3.5 & 2.7 & 2.1 & 0.0 & 13.0 & 20.2 & 19.0 & $\mathbf{20.7}$ \\
PGD ($\epsilon=0.075$) & 0.9 & 1.1 & 2.0 & 0.0 & 9.7 & 15.3 & $\mathbf{15.6}$ & 15.1 \\
\midrule
C\&W Perturb & 6.6 & 0.0 & 0.0 & 0.0 & 8.1 & 14.6 & $\mathbf{15.0}$ & 14.5 \\
\bottomrule
\end{tabular}
\end{adjustbox}
\caption{\textbf{Comparison of adversarial robustness in zero-shot 3D classification task.} The best scores are in bold. }
\label{tab:benchmark_adv}
\end{table*}

\vspace{-0.7cm}

\subsection{Ablation Study}

\begin{figure}[!htbp]
\centering
\subfigure[Noised point cloud with scale 0.02, 0.04, 0.06, 0.08, 0.10, and 0.12 from left to right (full time step).]
{
\begin{minipage}[t]{0.98\linewidth}
\centering
\includegraphics[width=0.98\linewidth, height=0.2\linewidth]{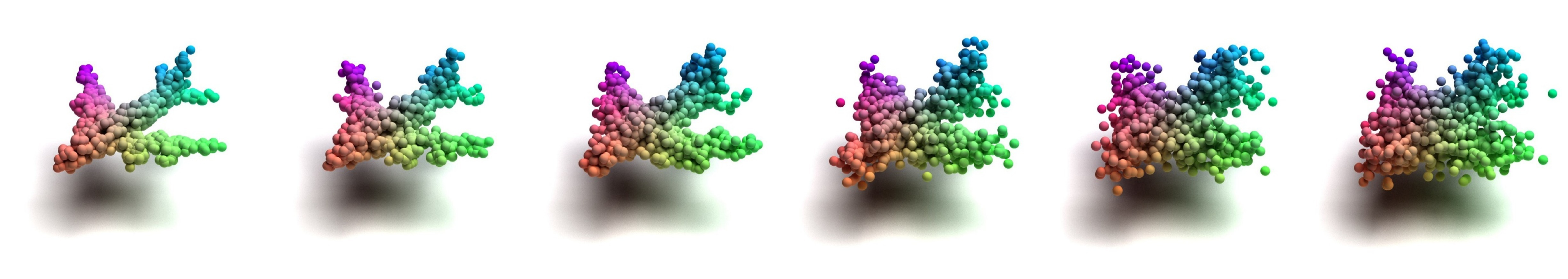}
\end{minipage}
}
\subfigure[Noised point cloud at time step 0, 200, 400, 600, 800, and 999 from left to right with scale 0.08.]
{
\begin{minipage}[t]{0.98\linewidth}
\centering
\includegraphics[width=0.98\linewidth, height=0.2\linewidth]{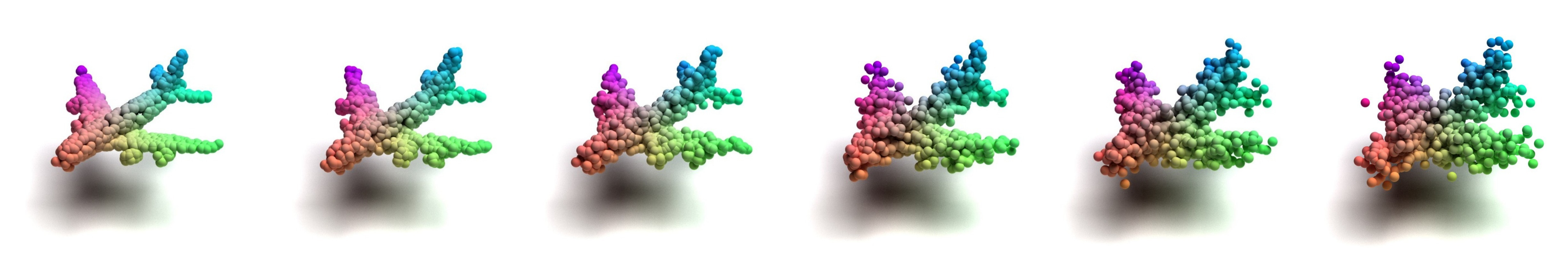}
\end{minipage}
}
\caption{\textbf{Visualization of different noise scaling factor (up) and noise time step (down).} }
\label{Fig:Point_Cloud_Visualization}
\end{figure}

\paragraph{Noise Scale.} Noise scale plays an important role in our method. If the noise scaling factor $s$ is too small, we can hardly learn robust 3D representation. If the factor is set too large, the pre-training would be difficult to converge. We first present Figure~\ref{Fig:Point_Cloud_Visualization} as a visualization for noise scale and noise time step on point cloud. In Figure~\ref{Fig:Point_Cloud_Visualization}(a), we change the noise scale and fix the time step to the max steps (999/1000). When $s$ is set to be too large like 0.10 or 0.12, even humans can hardly recognize it. When $s$ is set to be too small like 0.02, there is little difference from the clean data, as point cloud itself has contained some noise and perturbation. In Figure~\ref{Fig:Point_Cloud_Visualization}(b), we visualize the point cloud at different time steps with $s=0.08$. We can generally recognize the shape in most cases. We ablate $s$ in zero-shot classification task, as shown in Table~\ref{table:ablation_scaling_merging}(a). We find that even setting $s=0$ can still have a decent performance, \ie, forcing the network to predict the input point tokens themselves in the pre-training stage. This indicates that the most critical point for 3D self-supervised learning may not lie on the proxy task. This result is also similar to ReCon~\cite{qi2023contrast} (61.7\%), which also reveals the weak effect of mask reconstruction in this knowledge distillation setting.

\begin{table*}[!htbp]
\centering
\subtable[Ablation study on noise scaling factor $s$.]{
\resizebox{0.48\linewidth}{!}{
    \begin{tabular}{lccccccc}
    \toprule
        $s$ & 0 & 0.02 & 0.04 & 0.06 & 0.08 & 0.1 & 0.12 \\
        \midrule
        Acc & 60.4 & 65.6 & 67.3 & 68.1 & \textbf{69.0} & 68.3 & 67.9 \\
    \bottomrule
    \end{tabular}}
} 
\subtable[Ablation study on time step merging factor $\Delta$.]{
\resizebox{0.48\linewidth}{!}{
    \begin{tabular}{lccccccc}
    \toprule
        $\Delta$ & 1 & 10 & 20 & 50 & 100 & 200 & 1000 \\
        \midrule
        Acc & 66.9 & 67.5 & 67.3 & 68.1 & 68.7 & \textbf{69.0} & 65.4 \\
    \bottomrule
    \end{tabular}}
} 
\vspace{0.1cm}
\caption{\textbf{Ablation study on nosie scaling factor $s$ and time step merging factor $\Delta$ in zero-shot 3D classification task on ModelNet40 test set.} We report the best performance for each case.}
\label{table:ablation_scaling_merging}
\vspace{-0.1cm}
\end{table*}

\paragraph{Time Step Merging.} We ablate the time step merging interval parameter $\Delta$ in zero-shot 3D classification task. Results are shown in Table~\ref{table:ablation_scaling_merging}(b). Notice that $\Delta=1$ means do not conduct time step merging, and $\Delta=1000$ means do not use AdaLN-Zero in PointDAE. We can see that the time step embedding module plays a relatively important role for representation learning, and time step merging can slightly improve the performance of pre-training. 

\paragraph{Dual Denoising.} We ablate the two denoising design in our model, \ie, PointDAE and feature denoising. We set removing PointDAE as setting $s=0$, and set removing feature denoising as using learnable tokens to replace the standard gaussian noise for the input of feature branch. Results are shown in Table~\ref{table:ablation_denoising}. We can see that PointDAE and feature denoising both play a part for adversarial robustness, while PointDAE is more critical for both representation learning and adversarial robustness.

\begin{table*}[!htbp]
\centering
\subtable{
\resizebox{0.9\linewidth}{!}{
    \begin{tabular}{lccccccccccc}
    \toprule
        PGD Attack Step & 0 & 5 & 10 & 15 & 20 & 25 & 30 & 35 & 40 & 45 & 50 \\
        \midrule
        w/o. PointDAE & 60.4 & 41.1 & 38.3 & 37.8 & 36.9 & 35.1 & 34.2 & 33.9 & 33.7 & 33.5 & 32.8 \\
        w/o. Feature Denoising & 68.7 & 45.1 & 42.0 & 41.2 & 40.7 & 40.1 & 39.6 & 39.5 & 39.2 & 38.8 & 38.2 \\
        Complete & 69.0 & 49.3 & 43.7 & 42.1 & 41.5 & 40.9 & 40.6 & 40.4 & 40.3 & 40.2 & 39.9 \\
    \bottomrule
    \end{tabular}}
} 
\vspace{0.1cm}
\caption{\textbf{Ablation study on dual denoising in zero-shot 3D classification task on ModelNet40 test set.} We evaluate the robust accuracy under PGD attack with $\epsilon=0.01$. We report the performance for each model under the best inference configuration such as $N$ and $t$.}
\label{table:ablation_denoising}
\vspace{-0.1cm}
\end{table*}

\paragraph{Parallel Noise Inference and Noise Addition for Inference.} We conduct comprehensive experiments on the inference parameters $N$ and $t$ in Eq~\ref{eq:TTA} for both zero-shot classification and zero-shot adversarial robustness. Results are shown in Table~\ref{tab:Ablation_Parallel}. The difference of the distribution for ModelNet40 and ScanObjectNN makes the best configuration vary from each other, while we can generally conclude that using parallel noise inference and adding noise on the input point cloud can improve the performance and robustness in cross domain scenarios. We can also find that when $t=0$ and $N$ changes from 1 to 16, the performance can be significantly promoted, which shows the effectiveness of the feature denoising module. In this setting, the computation cost for inference does not increase significantly, because the point cloud feature extraction network (encoders in PointDAE) is repeatedly called and used as the cross attention context for the feature denoising branch, which does not require too much computation. Since we take the noised point cloud as the model input during pre-training, it is reasonable that adding some level of noise can slightly improve the performance, similar to $l$-DAE~\cite{chen2024deconstructing}.

\vspace{-0.4cm}
\begin{table}[!htbp]
\centering
\renewcommand\arraystretch{1.2}
{
\footnotesize
\begin{longtable}{ c | c | c  c  c  c  c  c  c  c  c  c }
\hline
& \diagbox[height=1.6\line]{${N}$}{${t}$} & {0} & {100} & {200} & {300} & {400} & {500} & {600} & {700} & {800} & {900} \\
\hline
\multirow{5}{*}{\begin{sideways}{\makecell[c]{{ModelNet40}}}\end{sideways}}
& {1} & 57.7 & 58.5 & 59.3 & 59.7 & 60.0 & 60.1 & 60.2 & 59.7 & 59.8 & 59.5 \\
& {2} & 61.5 & 62.8 & 63.0 & 63.1 & 63.5 & 63.6 & 63.3 & 62.9 & 62.7 & 62.7 \\
& {4} & 63.8 & 64.6 & 65.1 & 65.5 & 65.8 & 65.9 & 65.6 & 65.6 & 65.3 & 65.2 \\
& {8} & 66.5 & 67.3 & 68.1 & 68.6 & 68.8 & 68.9 & \textbf{69.0} & 68.9 & 68.9 & 68.8 \\
& {16} & 66.5 & 67.4 & 68.2 & 68.8 & 68.9 & \textbf{69.0} & \textbf{69.0} & \textbf{69.0} & 68.9 & 68.9 \\

\hline
\multirow{5}{*}{\begin{sideways}{\makecell[c]{{ModelNet40} \\ {Adversarial}}}\end{sideways}}
& {1} & 23.7 & 41.7 & 45.7 & \textbf{47.8} & 45.2 & 46.1 & 44.7 & 44.7 & 43.9 & 44.6 \\
& {2} & 21.7 & 39.5 & 44.7 & 46.6 & 44.2 & 45.5 & 45.2 & 45.5 & 44.7 & 44.1 \\
& {4} & 21.2 & 36.4 & 42.5 & 44.5 & 45.0 & 44.9 & 44.7 & 45.2 & 44.0 & 43.8 \\
& {8} & 21.7 & 30.3 & 38.8 & 41.3 & 42.9 & 43.4 & 43.2 & 42.3 & 42.0 & 42.2 \\
& {16} & 22.5 & 29.7 & 37.5 & 39.5 & 41.6 & 41.5 & 43.1 & 42.2 & 40.8 & 39.5 \\

\hline
\multirow{5}{*}{\begin{sideways}{\makecell[c]{{ScanObjNN}}}\end{sideways}}
& {1} & 44.4 & 42.7 & 41.1 & 40.3 & 41.0 & 40.8 & 41.0 & 40.4 & 41.0 & 40.1 \\
& {2} & 48.2 & 48.9 & 46.0 & 45.6 & 47.0 & 45.3 & 45.8 & 45.4 & 45.1 & 44.9 \\
& {4} & 49.9 & 50.4 & 49.1 & 49.2 & 48.2 & 47.5 & 47.2 & 46.6 & 47.5 & 47.3 \\
& {8} & 49.1 & \textbf{52.7} & 51.3 & 50.8 & 50.9 & 50.4 & 50.9 & 50.6 & 50.4 & 49.9 \\
& {16} & 49.1 & \textbf{52.7} & 51.6 & 50.8 & 50.9 & 50.6 & 51.1 & 50.8 & 50.4 & 50.4 \\
\hline

\multirow{5}{*}{\begin{sideways}{\makecell[c]{{ScanObjNN} \\ {Adversarial}}}\end{sideways}}
& {1} & 8.8 & 20.5 & 21.3 & 22.2 & 22.4 & 21.0 & 22.5 & 20.5 & 21.7 & 20.5 \\
& {2} & 10.0 & 23.1 & 27.0 & 25.6 & 22.7 & 23.9 & 23.8 & 24.1 & 24.4 & 25.0 \\
& {4} & 10.2 & 19.1 & 25.6 & 26.7 & 28.6 & 26.0 & 27.4 & \textbf{29.4} & 28.4 & 25.6 \\
& {8} & 10.7 & 15.5 & 27.7 & 25.6 & 25.3 & 28.2 & 27.4 & 29.1 & 26.7 & 25.1 \\
& {16} & 10.7 & 14.1 & 25.5 & 24.4 & 24.3 & 27.4 & 27.4 & 26.9 & 27.2 & 26.7 \\
\hline
\end{longtable}}
\vspace{0.2cm}
\caption{\textbf{Ablation study on $\boldsymbol{N}$ and $\boldsymbol{t}$ in zero-shot classification task on ModelNet40 and ScanObjNN test set.} Adversarial robustness is evaluated using PGD attack, optimizing for 10 steps with $\epsilon=0.01$.}
\label{tab:Ablation_Parallel}
\end{table}

\vspace{-0.4cm}
\section{Conclusion}\label{sec:conclusion}

In this paper, we propose \textit{Dual Denoising}, an algorithm to learn robust 3D representation from pre-trained VLMs like CLIP. Experiments on zero-shot recognition benchmark show that our method can generalize better than others with similar scale and settings, while experiments on zero-shot recognition under adversarial attack show that our method can learn more robust 3D representations from the proposed denoising design. We also propose to use parallel noise inference, \ie, using different gaussian noises with the same scale and average the output features. Experiments show that this design can significantly improve the cross domain performance, and also shows the effectiveness of the two modules we proposed, \ie, PointDAE and feature denoising. We would further explore to scale up our method in the following works to demonstrate whether \textit{Dual Denoising} is scalable for 3D representation learning. 

\newpage
\bibliography{neurips_2024}

\begin{thebibliography}{10}

\bibitem{radford2021learning}
Alec Radford, Jong~Wook Kim, Chris Hallacy, Aditya Ramesh, Gabriel Goh, Sandhini Agarwal, Girish Sastry, Amanda Askell, Pamela Mishkin, Jack Clark, et~al.
\newblock Learning transferable visual models from natural language supervision.
\newblock In {\em International conference on machine learning}, pages 8748--8763. PMLR, 2021.

\bibitem{Li_2023_CVPR_FLIP}
Yanghao Li, Haoqi Fan, Ronghang Hu, Christoph Feichtenhofer, and Kaiming He.
\newblock Scaling language-image pre-training via masking.
\newblock In {\em Proceedings of the IEEE/CVF Conference on Computer Vision and Pattern Recognition (CVPR)}, pages 23390--23400, June 2023.

\bibitem{mu2022slip}
Norman Mu, Alexander Kirillov, David Wagner, and Saining Xie.
\newblock Slip: Self-supervision meets language-image pre-training.
\newblock In {\em European conference on computer vision}, pages 529--544. Springer, 2022.

\bibitem{sun2023eva}
Quan Sun, Yuxin Fang, Ledell Wu, Xinlong Wang, and Yue Cao.
\newblock Eva-clip: Improved training techniques for clip at scale.
\newblock {\em arXiv preprint arXiv:2303.15389}, 2023.

\bibitem{zhang2022pointclip}
Renrui Zhang, Ziyu Guo, Wei Zhang, Kunchang Li, Xupeng Miao, Bin Cui, Yu~Qiao, Peng Gao, and Hongsheng Li.
\newblock Pointclip: Point cloud understanding by clip.
\newblock In {\em Proceedings of the IEEE/CVF conference on computer vision and pattern recognition}, pages 8552--8562, 2022.

\bibitem{zhu2023pointclip}
Xiangyang Zhu, Renrui Zhang, Bowei He, Ziyu Guo, Ziyao Zeng, Zipeng Qin, Shanghang Zhang, and Peng Gao.
\newblock Pointclip v2: Prompting clip and gpt for powerful 3d open-world learning.
\newblock In {\em Proceedings of the IEEE/CVF International Conference on Computer Vision}, pages 2639--2650, 2023.

\bibitem{Huang_2023_ICCV_CLIP2Point}
Tianyu Huang, Bowen Dong, Yunhan Yang, Xiaoshui Huang, Rynson~W.H. Lau, Wanli Ouyang, and Wangmeng Zuo.
\newblock Clip2point: Transfer clip to point cloud classification with image-depth pre-training.
\newblock In {\em Proceedings of the IEEE/CVF International Conference on Computer Vision (ICCV)}, pages 22157--22167, October 2023.

\bibitem{Zeng_2023_CVPR_CLIP2}
Yihan Zeng, Chenhan Jiang, Jiageng Mao, Jianhua Han, Chaoqiang Ye, Qingqiu Huang, Dit-Yan Yeung, Zhen Yang, Xiaodan Liang, and Hang Xu.
\newblock Clip2: Contrastive language-image-point pretraining from real-world point cloud data.
\newblock In {\em Proceedings of the IEEE/CVF Conference on Computer Vision and Pattern Recognition (CVPR)}, pages 15244--15253, June 2023.

\bibitem{Xue_2023_CVPR_ULIP}
Le~Xue, Mingfei Gao, Chen Xing, Roberto Mart{\'\i}n-Mart{\'\i}n, Jiajun Wu, Caiming Xiong, Ran Xu, Juan~Carlos Niebles, and Silvio Savarese.
\newblock Ulip: Learning a unified representation of language, images, and point clouds for 3d understanding.
\newblock In {\em Proceedings of the IEEE/CVF Conference on Computer Vision and Pattern Recognition (CVPR)}, pages 1179--1189, June 2023.

\bibitem{qi2023contrast}
Zekun Qi, Runpei Dong, Guofan Fan, Zheng Ge, Xiangyu Zhang, Kaisheng Ma, and Li~Yi.
\newblock Contrast with reconstruct: Contrastive 3d representation learning guided by generative pretraining.
\newblock In {\em International Conference on Machine Learning}, pages 28223--28243. PMLR, 2023.

\bibitem{zhou2023uni3d}
Junsheng Zhou, Jinsheng Wang, Baorui Ma, Yu-Shen Liu, Tiejun Huang, and Xinlong Wang.
\newblock Uni3d: Exploring unified 3d representation at scale.
\newblock In {\em The Twelfth International Conference on Learning Representations}, 2023.

\bibitem{xue2023ulip_2}
Le~Xue, Ning Yu, Shu Zhang, Artemis Panagopoulou, Junnan Li, Roberto Mart{\'\i}n-Mart{\'\i}n, Jiajun Wu, Caiming Xiong, Ran Xu, Juan~Carlos Niebles, et~al.
\newblock Ulip-2: Towards scalable multimodal pre-training for 3d understanding.
\newblock {\em arXiv preprint arXiv:2305.08275}, 2023.

\bibitem{shafahi2019adversarial}
Ali Shafahi, Mahyar Najibi, Mohammad~Amin Ghiasi, Zheng Xu, John Dickerson, Christoph Studer, Larry~S Davis, Gavin Taylor, and Tom Goldstein.
\newblock Adversarial training for free!
\newblock {\em Advances in neural information processing systems}, 32, 2019.

\bibitem{xie2019feature}
Cihang Xie, Yuxin Wu, Laurens van~der Maaten, Alan~L Yuille, and Kaiming He.
\newblock Feature denoising for improving adversarial robustness.
\newblock In {\em Proceedings of the IEEE/CVF conference on computer vision and pattern recognition}, pages 501--509, 2019.

\bibitem{yoon2021adversarial}
Jongmin Yoon, Sung~Ju Hwang, and Juho Lee.
\newblock Adversarial purification with score-based generative models.
\newblock In {\em International Conference on Machine Learning}, pages 12062--12072. PMLR, 2021.

\bibitem{nie2022diffusion}
Weili Nie, Brandon Guo, Yujia Huang, Chaowei Xiao, Arash Vahdat, and Animashree Anandkumar.
\newblock Diffusion models for adversarial purification.
\newblock In {\em International Conference on Machine Learning}, pages 16805--16827. PMLR, 2022.

\bibitem{vincent2008extracting}
Pascal Vincent, Hugo Larochelle, Yoshua Bengio, and Pierre-Antoine Manzagol.
\newblock Extracting and composing robust features with denoising autoencoders.
\newblock In {\em Proceedings of the 25th international conference on Machine learning}, pages 1096--1103, 2008.

\bibitem{ho2020denoising}
Jonathan Ho, Ajay Jain, and Pieter Abbeel.
\newblock Denoising diffusion probabilistic models.
\newblock {\em Advances in neural information processing systems}, 33:6840--6851, 2020.

\bibitem{chen2024deconstructing}
Xinlei Chen, Zhuang Liu, Saining Xie, and Kaiming He.
\newblock Deconstructing denoising diffusion models for self-supervised learning.
\newblock {\em arXiv preprint arXiv:2401.14404}, 2024.

\bibitem{zhai2023sigmoid}
Xiaohua Zhai, Basil Mustafa, Alexander Kolesnikov, and Lucas Beyer.
\newblock Sigmoid loss for language image pre-training.
\newblock In {\em Proceedings of the IEEE/CVF International Conference on Computer Vision}, pages 11975--11986, 2023.

\bibitem{qi2017pointnet}
Charles~R Qi, Hao Su, Kaichun Mo, and Leonidas~J Guibas.
\newblock Pointnet: Deep learning on point sets for 3d classification and segmentation.
\newblock In {\em Proceedings of the IEEE conference on computer vision and pattern recognition}, pages 652--660, 2017.

\bibitem{qi2017pointnet++}
Charles~Ruizhongtai Qi, Li~Yi, Hao Su, and Leonidas~J Guibas.
\newblock Pointnet++: Deep hierarchical feature learning on point sets in a metric space.
\newblock {\em Advances in neural information processing systems}, 30, 2017.

\bibitem{guo2021pct}
Meng-Hao Guo, Jun-Xiong Cai, Zheng-Ning Liu, Tai-Jiang Mu, Ralph~R Martin, and Shi-Min Hu.
\newblock Pct: Point cloud transformer.
\newblock {\em Computational Visual Media}, 7:187--199, 2021.

\bibitem{zhao2021point}
Hengshuang Zhao, Li~Jiang, Jiaya Jia, Philip~HS Torr, and Vladlen Koltun.
\newblock Point transformer.
\newblock In {\em Proceedings of the IEEE/CVF international conference on computer vision}, pages 16259--16268, 2021.

\bibitem{goodfellow2014explaining}
Ian~J Goodfellow, Jonathon Shlens, and Christian Szegedy.
\newblock Explaining and harnessing adversarial examples.
\newblock {\em arXiv preprint arXiv:1412.6572}, 2014.

\bibitem{madry2018towards}
Aleksander Madry, Aleksandar Makelov, Ludwig Schmidt, Dimitris Tsipras, and Adrian Vladu.
\newblock Towards deep learning models resistant to adversarial attacks.
\newblock In {\em International Conference on Learning Representations}, 2018.

\bibitem{xiang2019generating}
Chong Xiang, Charles~R Qi, and Bo~Li.
\newblock Generating 3d adversarial point clouds.
\newblock In {\em Proceedings of the IEEE/CVF conference on computer vision and pattern recognition}, pages 9136--9144, 2019.

\bibitem{hamdi2020advpc}
Abdullah Hamdi, Sara Rojas, Ali Thabet, and Bernard Ghanem.
\newblock Advpc: Transferable adversarial perturbations on 3d point clouds.
\newblock In {\em Computer Vision--ECCV 2020: 16th European Conference, Glasgow, UK, August 23--28, 2020, Proceedings, Part XII 16}, pages 241--257. Springer, 2020.

\bibitem{zhou2019dup}
Hang Zhou, Kejiang Chen, Weiming Zhang, Han Fang, Wenbo Zhou, and Nenghai Yu.
\newblock Dup-net: Denoiser and upsampler network for 3d adversarial point clouds defense.
\newblock In {\em Proceedings of the IEEE/CVF international conference on computer vision}, pages 1961--1970, 2019.

\bibitem{sun2020adversarial}
Jiachen Sun, Karl Koenig, Yulong Cao, Qi~Alfred Chen, and Z~Morley Mao.
\newblock On adversarial robustness of 3d point cloud classification under adaptive attacks.
\newblock {\em arXiv preprint arXiv:2011.11922}, 2020.

\bibitem{sun2023critical}
Jiachen Sun, Jiongxiao Wang, Weili Nie, Zhiding Yu, Zhuoqing Mao, and Chaowei Xiao.
\newblock A critical revisit of adversarial robustness in 3d point cloud recognition with diffusion-driven purification.
\newblock In {\em International Conference on Machine Learning}, pages 33100--33114. PMLR, 2023.

\bibitem{dhariwal2021diffusion}
Prafulla Dhariwal and Alexander Nichol.
\newblock Diffusion models beat gans on image synthesis.
\newblock {\em Advances in neural information processing systems}, 34:8780--8794, 2021.

\bibitem{rombach2022high}
Robin Rombach, Andreas Blattmann, Dominik Lorenz, Patrick Esser, and Bj{\"o}rn Ommer.
\newblock High-resolution image synthesis with latent diffusion models.
\newblock In {\em Proceedings of the IEEE/CVF conference on computer vision and pattern recognition}, pages 10684--10695, 2022.

\bibitem{peebles2023scalable}
William Peebles and Saining Xie.
\newblock Scalable diffusion models with transformers.
\newblock In {\em Proceedings of the IEEE/CVF International Conference on Computer Vision}, pages 4195--4205, 2023.

\bibitem{xiang2023denoising}
Weilai Xiang, Hongyu Yang, Di~Huang, and Yunhong Wang.
\newblock Denoising diffusion autoencoders are unified self-supervised learners.
\newblock In {\em Proceedings of the IEEE/CVF International Conference on Computer Vision}, pages 15802--15812, 2023.

\bibitem{chen2021exploring}
Xinlei Chen and Kaiming He.
\newblock Exploring simple siamese representation learning.
\newblock In {\em Proceedings of the IEEE/CVF conference on computer vision and pattern recognition}, pages 15750--15758, 2021.

\bibitem{dosovitskiy2020image}
Alexey Dosovitskiy, Lucas Beyer, Alexander Kolesnikov, Dirk Weissenborn, Xiaohua Zhai, Thomas Unterthiner, Mostafa Dehghani, Matthias Minderer, Georg Heigold, Sylvain Gelly, et~al.
\newblock An image is worth 16x16 words: Transformers for image recognition at scale.
\newblock {\em arXiv preprint arXiv:2010.11929}, 2020.

\bibitem{chang2015shapenet}
Angel~X Chang, Thomas Funkhouser, Leonidas Guibas, Pat Hanrahan, Qixing Huang, Zimo Li, Silvio Savarese, Manolis Savva, Shuran Song, Hao Su, et~al.
\newblock Shapenet: An information-rich 3d model repository.
\newblock {\em arXiv preprint arXiv:1512.03012}, 2015.

\bibitem{deitke2023objaverse}
Matt Deitke, Dustin Schwenk, Jordi Salvador, Luca Weihs, Oscar Michel, Eli VanderBilt, Ludwig Schmidt, Kiana Ehsani, Aniruddha Kembhavi, and Ali Farhadi.
\newblock Objaverse: A universe of annotated 3d objects.
\newblock In {\em Proceedings of the IEEE/CVF Conference on Computer Vision and Pattern Recognition}, pages 13142--13153, 2023.

\bibitem{qi2024shapellm}
Zekun Qi, Runpei Dong, Shaochen Zhang, Haoran Geng, Chunrui Han, Zheng Ge, He~Wang, Li~Yi, and Kaisheng Ma.
\newblock Shapellm: Universal 3d object understanding for embodied interaction.
\newblock {\em arXiv preprint arXiv:2402.17766}, 2024.

\bibitem{vaswani2017attention}
Ashish Vaswani, Noam Shazeer, Niki Parmar, Jakob Uszkoreit, Llion Jones, Aidan~N Gomez, {\L}ukasz Kaiser, and Illia Polosukhin.
\newblock Attention is all you need.
\newblock {\em Advances in neural information processing systems}, 30, 2017.

\bibitem{uy2019revisiting}
Mikaela~Angelina Uy, Quang-Hieu Pham, Binh-Son Hua, Thanh Nguyen, and Sai-Kit Yeung.
\newblock Revisiting point cloud classification: A new benchmark dataset and classification model on real-world data.
\newblock In {\em Proceedings of the IEEE/CVF international conference on computer vision}, pages 1588--1597, 2019.

\bibitem{wu20153d}
Zhirong Wu, Shuran Song, Aditya Khosla, Fisher Yu, Linguang Zhang, Xiaoou Tang, and Jianxiong Xiao.
\newblock 3d shapenets: A deep representation for volumetric shapes.
\newblock In {\em Proceedings of the IEEE conference on computer vision and pattern recognition}, pages 1912--1920, 2015.

\bibitem{ding2023cap}
Daizong Ding, Erling Jiang, Yuanmin Huang, Mi~Zhang, Wenxuan Li, and Min Yang.
\newblock Cap: Robust point cloud classification via semantic and structural modeling.
\newblock In {\em Proceedings of the IEEE/CVF Conference on Computer Vision and Pattern Recognition}, pages 12260--12270, 2023.

\bibitem{sun2021adversarially}
Jiachen Sun, Yulong Cao, Christopher~B Choy, Zhiding Yu, Anima Anandkumar, Zhuoqing~Morley Mao, and Chaowei Xiao.
\newblock Adversarially robust 3d point cloud recognition using self-supervisions.
\newblock {\em Advances in Neural Information Processing Systems}, 34:15498--15512, 2021.

\end{thebibliography}
\bibliographystyle{unsrt}


\appendix

\section{Details for Adversarial Attack on Point Cloud}\label{appendix}

We use IFGM, PGD and C\&W attack in this paper, where C\&W attack is $L_2$ norm-based and the others are $L_{\infty}$ norm based. For C\&W attack, we set the loss function as:
\begin{equation}
    \mathcal{L}=(\max\limits_{i\neq t^{\prime}}\mathcal{Z}(\boldsymbol{X}^{\prime})_{i}-\mathcal{Z}(\boldsymbol{X}^{\prime})_{t^{\prime}})^{+}+\lambda\cdot\lVert \boldsymbol{X}-\boldsymbol{X}^{\prime} \rVert_2,
\end{equation}
where $\mathbf{X}\in\mathbb{R}^{n\times 3}$ is the clean point cloud, $\mathbf{X}^{\prime}\in\mathbb{R}^{n\times 3}$ is the optimized adversarial point cloud, $\mathcal{Z}(\boldsymbol{X})_i$ is the $i$-th element of the output logits, and $t^{\prime}$ is the target class. Here logits are computed via dot-production between the point cloud feature and the set of text CLIP features. We leverage 10-step binary search to find the appropriate hyper parameter $\lambda$ from [10, 80]. We use the whole test set of ModelNet40 and ScanObjectNN (OBJ\_ONLY) for evaluation. The step size of the adversarial optimization is 0.01 and we allow at most 500 iterations of optimization in each binary search to find the adversarial examples. For the $L_{\infty}$ norm-based PGD attack, we adopt the formulation as:
\begin{equation}
    \boldsymbol{X}_{t+1}=\Pi_{\boldsymbol{X}+\mathcal{S}}(\boldsymbol{X}_t+\alpha\cdot\text{sign}(\nabla_{\boldsymbol{X}_t}\mathcal{L}(\boldsymbol{X}_t, \boldsymbol{\theta}, \boldsymbol{y}))),
\end{equation}
where $\boldsymbol{X}_t$ is the adversarial point cloud in the $t$-th iteration during attack, $\Pi$ is the projection function to project the adversarial point cloud to a pre-defined space $\boldsymbol{X}+\mathcal{S}$, the $L_{\infty}$ norm ball. $\alpha$ is the step size. We use the $\text{sign}$ function to normalize the gradient into $L_{\infty}$ norm ball at each iteration. We set the boundary of allowed perturbations as $\epsilon=\{0.01, 0.025, 0.05, 0.075\}$ for space $\mathcal{S}$. Since point cloud data is continuous within the range of $[-1,1]$, we set the step size as $\alpha=\epsilon/10$. IFGM is basically similar to PGD, with a difference in perturbation initialization.


\newpage
\section*{NeurIPS Paper Checklist}

\begin{enumerate}

\item {\bf Claims}
    \item[] Question: Do the main claims made in the abstract and introduction accurately reflect the paper's contributions and scope?
    \item[] Answer: \answerYes{} 
    \item[] Justification: We claim the paper's contributions and scope in the abstract and introduction.
    \item[] Guidelines:
    \begin{itemize}
        \item The answer NA means that the abstract and introduction do not include the claims made in the paper.
        \item The abstract and/or introduction should clearly state the claims made, including the contributions made in the paper and important assumptions and limitations. A No or NA answer to this question will not be perceived well by the reviewers. 
        \item The claims made should match theoretical and experimental results, and reflect how much the results can be expected to generalize to other settings. 
        \item It is fine to include aspirational goals as motivation as long as it is clear that these goals are not attained by the paper. 
    \end{itemize}

\item {\bf Limitations}
    \item[] Question: Does the paper discuss the limitations of the work performed by the authors?
    \item[] Answer: \answerYes{} 
    \item[] Justification: We discuss the limitations in Section~\ref{sec:conclusion}, which is we do not explore to scale up our method in this paper.
    \item[] Guidelines:
    \begin{itemize}
        \item The answer NA means that the paper has no limitation while the answer No means that the paper has limitations, but those are not discussed in the paper. 
        \item The authors are encouraged to create a separate "Limitations" section in their paper.
        \item The paper should point out any strong assumptions and how robust the results are to violations of these assumptions (e.g., independence assumptions, noiseless settings, model well-specification, asymptotic approximations only holding locally). The authors should reflect on how these assumptions might be violated in practice and what the implications would be.
        \item The authors should reflect on the scope of the claims made, e.g., if the approach was only tested on a few datasets or with a few runs. In general, empirical results often depend on implicit assumptions, which should be articulated.
        \item The authors should reflect on the factors that influence the performance of the approach. For example, a facial recognition algorithm may perform poorly when image resolution is low or images are taken in low lighting. Or a speech-to-text system might not be used reliably to provide closed captions for online lectures because it fails to handle technical jargon.
        \item The authors should discuss the computational efficiency of the proposed algorithms and how they scale with dataset size.
        \item If applicable, the authors should discuss possible limitations of their approach to address problems of privacy and fairness.
        \item While the authors might fear that complete honesty about limitations might be used by reviewers as grounds for rejection, a worse outcome might be that reviewers discover limitations that aren't acknowledged in the paper. The authors should use their best judgment and recognize that individual actions in favor of transparency play an important role in developing norms that preserve the integrity of the community. Reviewers will be specifically instructed to not penalize honesty concerning limitations.
    \end{itemize}

\item {\bf Theory Assumptions and Proofs}
    \item[] Question: For each theoretical result, does the paper provide the full set of assumptions and a complete (and correct) proof?
    \item[] Answer: \answerNA{} 
    \item[] Justification: Our paper does not include theoretical results.
    \item[] Guidelines:
    \begin{itemize}
        \item The answer NA means that the paper does not include theoretical results. 
        \item All the theorems, formulas, and proofs in the paper should be numbered and cross-referenced.
        \item All assumptions should be clearly stated or referenced in the statement of any theorems.
        \item The proofs can either appear in the main paper or the supplemental material, but if they appear in the supplemental material, the authors are encouraged to provide a short proof sketch to provide intuition. 
        \item Inversely, any informal proof provided in the core of the paper should be complemented by formal proofs provided in appendix or supplemental material.
        \item Theorems and Lemmas that the proof relies upon should be properly referenced. 
    \end{itemize}

    \item {\bf Experimental Result Reproducibility}
    \item[] Question: Does the paper fully disclose all the information needed to reproduce the main experimental results of the paper to the extent that it affects the main claims and/or conclusions of the paper (regardless of whether the code and data are provided or not)?
    \item[] Answer: \answerYes{} 
    \item[] Justification: We fully described the key points of the implementation details so that it can be reproduced by other researchers.
    \item[] Guidelines:
    \begin{itemize}
        \item The answer NA means that the paper does not include experiments.
        \item If the paper includes experiments, a No answer to this question will not be perceived well by the reviewers: Making the paper reproducible is important, regardless of whether the code and data are provided or not.
        \item If the contribution is a dataset and/or model, the authors should describe the steps taken to make their results reproducible or verifiable. 
        \item Depending on the contribution, reproducibility can be accomplished in various ways. For example, if the contribution is a novel architecture, describing the architecture fully might suffice, or if the contribution is a specific model and empirical evaluation, it may be necessary to either make it possible for others to replicate the model with the same dataset, or provide access to the model. In general. releasing code and data is often one good way to accomplish this, but reproducibility can also be provided via detailed instructions for how to replicate the results, access to a hosted model (e.g., in the case of a large language model), releasing of a model checkpoint, or other means that are appropriate to the research performed.
        \item While NeurIPS does not require releasing code, the conference does require all submissions to provide some reasonable avenue for reproducibility, which may depend on the nature of the contribution. For example
        \begin{enumerate}
            \item If the contribution is primarily a new algorithm, the paper should make it clear how to reproduce that algorithm.
            \item If the contribution is primarily a new model architecture, the paper should describe the architecture clearly and fully.
            \item If the contribution is a new model (e.g., a large language model), then there should either be a way to access this model for reproducing the results or a way to reproduce the model (e.g., with an open-source dataset or instructions for how to construct the dataset).
            \item We recognize that reproducibility may be tricky in some cases, in which case authors are welcome to describe the particular way they provide for reproducibility. In the case of closed-source models, it may be that access to the model is limited in some way (e.g., to registered users), but it should be possible for other researchers to have some path to reproducing or verifying the results.
        \end{enumerate}
    \end{itemize}

\item {\bf Open access to data and code}
    \item[] Question: Does the paper provide open access to the data and code, with sufficient instructions to faithfully reproduce the main experimental results, as described in supplemental material?
    \item[] Answer: \answerYes{} 
    \item[] Justification: The dataset we used in our paper is openly accessible, and we provide the source code in the supplementary material.
    \item[] Guidelines:
    \begin{itemize}
        \item The answer NA means that paper does not include experiments requiring code.
        \item Please see the NeurIPS code and data submission guidelines (\url{https://nips.cc/public/guides/CodeSubmissionPolicy}) for more details.
        \item While we encourage the release of code and data, we understand that this might not be possible, so “No” is an acceptable answer. Papers cannot be rejected simply for not including code, unless this is central to the contribution (e.g., for a new open-source benchmark).
        \item The instructions should contain the exact command and environment needed to run to reproduce the results. See the NeurIPS code and data submission guidelines (\url{https://nips.cc/public/guides/CodeSubmissionPolicy}) for more details.
        \item The authors should provide instructions on data access and preparation, including how to access the raw data, preprocessed data, intermediate data, and generated data, etc.
        \item The authors should provide scripts to reproduce all experimental results for the new proposed method and baselines. If only a subset of experiments are reproducible, they should state which ones are omitted from the script and why.
        \item At submission time, to preserve anonymity, the authors should release anonymized versions (if applicable).
        \item Providing as much information as possible in supplemental material (appended to the paper) is recommended, but including URLs to data and code is permitted.
    \end{itemize}

\item {\bf Experimental Setting/Details}
    \item[] Question: Does the paper specify all the training and test details (e.g., data splits, hyperparameters, how they were chosen, type of optimizer, etc.) necessary to understand the results?
    \item[] Answer: \answerYes{} 
    \item[] Justification: We present the necessary experiment settings in our paper.
    \item[] Guidelines:
    \begin{itemize}
        \item The answer NA means that the paper does not include experiments.
        \item The experimental setting should be presented in the core of the paper to a level of detail that is necessary to appreciate the results and make sense of them.
        \item The full details can be provided either with the code, in appendix, or as supplemental material.
    \end{itemize}

\item {\bf Experiment Statistical Significance}
    \item[] Question: Does the paper report error bars suitably and correctly defined or other appropriate information about the statistical significance of the experiments?
    \item[] Answer: \answerNo{} 
    \item[] Justification: It is unnecessary to do this on the benchmark used in this paper.
    \item[] Guidelines:
    \begin{itemize}
        \item The answer NA means that the paper does not include experiments.
        \item The authors should answer "Yes" if the results are accompanied by error bars, confidence intervals, or statistical significance tests, at least for the experiments that support the main claims of the paper.
        \item The factors of variability that the error bars are capturing should be clearly stated (for example, train/test split, initialization, random drawing of some parameter, or overall run with given experimental conditions).
        \item The method for calculating the error bars should be explained (closed form formula, call to a library function, bootstrap, etc.)
        \item The assumptions made should be given (e.g., Normally distributed errors).
        \item It should be clear whether the error bar is the standard deviation or the standard error of the mean.
        \item It is OK to report 1-sigma error bars, but one should state it. The authors should preferably report a 2-sigma error bar than state that they have a 96\% CI, if the hypothesis of Normality of errors is not verified.
        \item For asymmetric distributions, the authors should be careful not to show in tables or figures symmetric error bars that would yield results that are out of range (e.g. negative error rates).
        \item If error bars are reported in tables or plots, The authors should explain in the text how they were calculated and reference the corresponding figures or tables in the text.
    \end{itemize}

\item {\bf Experiments Compute Resources}
    \item[] Question: For each experiment, does the paper provide sufficient information on the computer resources (type of compute workers, memory, time of execution) needed to reproduce the experiments?
    \item[] Answer: \answerYes{} 
    \item[] Justification: We present the computer resources needed for experiment implementation.
    \item[] Guidelines:
    \begin{itemize}
        \item The answer NA means that the paper does not include experiments.
        \item The paper should indicate the type of compute workers CPU or GPU, internal cluster, or cloud provider, including relevant memory and storage.
        \item The paper should provide the amount of compute required for each of the individual experimental runs as well as estimate the total compute. 
        \item The paper should disclose whether the full research project required more compute than the experiments reported in the paper (e.g., preliminary or failed experiments that didn't make it into the paper). 
    \end{itemize}
    
\item {\bf Code Of Ethics}
    \item[] Question: Does the research conducted in the paper conform, in every respect, with the NeurIPS Code of Ethics \url{https://neurips.cc/public/EthicsGuidelines}?
    \item[] Answer: \answerYes{} 
    \item[] Justification: We conform the code of ethics.
    \item[] Guidelines:
    \begin{itemize}
        \item The answer NA means that the authors have not reviewed the NeurIPS Code of Ethics.
        \item If the authors answer No, they should explain the special circumstances that require a deviation from the Code of Ethics.
        \item The authors should make sure to preserve anonymity (e.g., if there is a special consideration due to laws or regulations in their jurisdiction).
    \end{itemize}

\item {\bf Broader Impacts}
    \item[] Question: Does the paper discuss both potential positive societal impacts and negative societal impacts of the work performed?
    \item[] Answer: \answerNA{} 
    \item[] Justification: Our work is purely a research paper.
    \item[] Guidelines:
    \begin{itemize}
        \item The answer NA means that there is no societal impact of the work performed.
        \item If the authors answer NA or No, they should explain why their work has no societal impact or why the paper does not address societal impact.
        \item Examples of negative societal impacts include potential malicious or unintended uses (e.g., disinformation, generating fake profiles, surveillance), fairness considerations (e.g., deployment of technologies that could make decisions that unfairly impact specific groups), privacy considerations, and security considerations.
        \item The conference expects that many papers will be foundational research and not tied to particular applications, let alone deployments. However, if there is a direct path to any negative applications, the authors should point it out. For example, it is legitimate to point out that an improvement in the quality of generative models could be used to generate deepfakes for disinformation. On the other hand, it is not needed to point out that a generic algorithm for optimizing neural networks could enable people to train models that generate Deepfakes faster.
        \item The authors should consider possible harms that could arise when the technology is being used as intended and functioning correctly, harms that could arise when the technology is being used as intended but gives incorrect results, and harms following from (intentional or unintentional) misuse of the technology.
        \item If there are negative societal impacts, the authors could also discuss possible mitigation strategies (e.g., gated release of models, providing defenses in addition to attacks, mechanisms for monitoring misuse, mechanisms to monitor how a system learns from feedback over time, improving the efficiency and accessibility of ML).
    \end{itemize}
    
\item {\bf Safeguards}
    \item[] Question: Does the paper describe safeguards that have been put in place for responsible release of data or models that have a high risk for misuse (e.g., pre-trained language models, image generators, or scraped datasets)?
    \item[] Answer: \answerNA{} 
    \item[] Justification: The paper poses no such risks.
    \item[] Guidelines:
    \begin{itemize}
        \item The answer NA means that the paper poses no such risks.
        \item Released models that have a high risk for misuse or dual-use should be released with necessary safeguards to allow for controlled use of the model, for example by requiring that users adhere to usage guidelines or restrictions to access the model or implementing safety filters. 
        \item Datasets that have been scraped from the Internet could pose safety risks. The authors should describe how they avoided releasing unsafe images.
        \item We recognize that providing effective safeguards is challenging, and many papers do not require this, but we encourage authors to take this into account and make a best faith effort.
    \end{itemize}

\item {\bf Licenses for existing assets}
    \item[] Question: Are the creators or original owners of assets (e.g., code, data, models), used in the paper, properly credited and are the license and terms of use explicitly mentioned and properly respected?
    \item[] Answer: \answerYes{} 
    \item[] Justification: We followed this guidance.
    \item[] Guidelines:
    \begin{itemize}
        \item The answer NA means that the paper does not use existing assets.
        \item The authors should cite the original paper that produced the code package or dataset.
        \item The authors should state which version of the asset is used and, if possible, include a URL.
        \item The name of the license (e.g., CC-BY 4.0) should be included for each asset.
        \item For scraped data from a particular source (e.g., website), the copyright and terms of service of that source should be provided.
        \item If assets are released, the license, copyright information, and terms of use in the package should be provided. For popular datasets, \url{paperswithcode.com/datasets} has curated licenses for some datasets. Their licensing guide can help determine the license of a dataset.
        \item For existing datasets that are re-packaged, both the original license and the license of the derived asset (if it has changed) should be provided.
        \item If this information is not available online, the authors are encouraged to reach out to the asset's creators.
    \end{itemize}

\item {\bf New Assets}
    \item[] Question: Are new assets introduced in the paper well documented and is the documentation provided alongside the assets?
    \item[] Answer: \answerNA{} 
    \item[] Justification: We do not release new assets.
    \item[] Guidelines:
    \begin{itemize}
        \item The answer NA means that the paper does not release new assets.
        \item Researchers should communicate the details of the dataset/code/model as part of their submissions via structured templates. This includes details about training, license, limitations, etc. 
        \item The paper should discuss whether and how consent was obtained from people whose asset is used.
        \item At submission time, remember to anonymize your assets (if applicable). You can either create an anonymized URL or include an anonymized zip file.
    \end{itemize}

\item {\bf Crowdsourcing and Research with Human Subjects}
    \item[] Question: For crowdsourcing experiments and research with human subjects, does the paper include the full text of instructions given to participants and screenshots, if applicable, as well as details about compensation (if any)? 
    \item[] Answer: \answerNA{} 
    \item[] Justification: This paper does not involve crowdsourcing nor research with human subjects.
    \item[] Guidelines:
    \begin{itemize}
        \item The answer NA means that the paper does not involve crowdsourcing nor research with human subjects.
        \item Including this information in the supplemental material is fine, but if the main contribution of the paper involves human subjects, then as much detail as possible should be included in the main paper. 
        \item According to the NeurIPS Code of Ethics, workers involved in data collection, curation, or other labor should be paid at least the minimum wage in the country of the data collector. 
    \end{itemize}

\item {\bf Institutional Review Board (IRB) Approvals or Equivalent for Research with Human Subjects}
    \item[] Question: Does the paper describe potential risks incurred by study participants, whether such risks were disclosed to the subjects, and whether Institutional Review Board (IRB) approvals (or an equivalent approval/review based on the requirements of your country or institution) were obtained?
    \item[] Answer: \answerNA{} 
    \item[] Justification: This paper does not involve crowdsourcing nor research with human subjects.
    \item[] Guidelines:
    \begin{itemize}
        \item The answer NA means that the paper does not involve crowdsourcing nor research with human subjects.
        \item Depending on the country in which research is conducted, IRB approval (or equivalent) may be required for any human subjects research. If you obtained IRB approval, you should clearly state this in the paper. 
        \item We recognize that the procedures for this may vary significantly between institutions and locations, and we expect authors to adhere to the NeurIPS Code of Ethics and the guidelines for their institution. 
        \item For initial submissions, do not include any information that would break anonymity (if applicable), such as the institution conducting the review.
    \end{itemize}

\end{enumerate}

\end{document}